\documentclass{article}
\usepackage{spconf,amsmath,graphicx,titleforsup}
\usepackage{amssymb}
\usepackage{amsfonts}
\usepackage{algorithmic}
\usepackage{algorithm}
\usepackage{multirow}
\usepackage{array}
\usepackage{url}
\usepackage{fancyhdr}
 
\fancypagestyle{firstpagefooter}{
  \fancyhf{}  
  \fancyfoot[C]{\scriptsize
    © 2025 IEEE. Personal use of this material is permitted. 
    Permission from IEEE must be obtained for all other uses, in any current or future media, 
    including reprinting/republishing this material for advertising or promotional purposes, 
    creating new collective works, for resale or redistribution to servers or lists, 
    or reuse of any copyrighted component of this work in other works.
  }

}

\title{CURVE: CLIP-Utilized Reinforcement learning for Visual image Enhancement via Simple Image Processing}

\name{Yuka Ogino, Takahiro Toizumi, and Atsushi Ito}
\address{NEC Corporation}

\begin{document}
%
\maketitle
\thispagestyle{firstpagefooter}
\begin{abstract}
Low-Light Image Enhancement (LLIE) is crucial for improving both human perception and computer vision tasks. This paper addresses two challenges in zero-reference LLIE: obtaining perceptually 'good' images using the Contrastive Language-Image Pre-Training (CLIP) model and maintaining computational efficiency for high-resolution images. We propose CLIP-Utilized Reinforcement learning-based Visual image Enhancement (CURVE). CURVE employs a simple image processing module which adjusts global image tone based on Bézier curve and estimates its processing parameters iteratively. The estimator is trained by reinforcement learning with rewards designed using CLIP text embeddings. Experiments on low-light and multi-exposure datasets demonstrate the performance of CURVE in terms of enhancement quality and processing speed compared to conventional methods.

\end{abstract}
\begin{keywords}
Image enhancement, image processing, reinforcement learning
\end{keywords}

\section{Introduction}
\label{sec:intro}
Lighting conditions, especially low-light scenarios, degrade image contrast. This degradation impacts both human perception and performance of computer vision tasks such as image recognition and object detection. To address these problems, Low-Light Image Enhancement (LLIE) methods have been proposed. Recent research has focused on zero-reference learning-based LLIE methods that do not require paired low-light and well-lit images. Some approaches \cite{Liu_2021_CVPR,Ma_2022_CVPR} apply models based on Retinex theory, while others, such as Zero-DCE \cite{Guo_2020_CVPR}, estimate tone curve adjustment parameters for each pixel using Convolutional Neural Networks (CNNs). These methods employ unsupervised loss functions based on statistical values and structural similarities of images. However, the effectiveness of these loss functions in producing perceptually preferable results is still being explored. 

Recently, Contrastive Language-Image Pre- Training (CLIP) \cite{clip} has emerged as a powerful tool in computer vision tasks. CLIP learns an embedding space that connects images and texts from large-scale image-text data. This capability has led to recent image enhancement techniques \cite{Liang_2023_ICCV,Morawski_2024_CVPR,morawski2024}. These CLIP-based image enhancement methods follow a two-step process. First, they optimize negative and positive prompts in the CLIP embedding space using poorly-lit and well-lit image groups. Then, they train the image enhancement model using the similarity between these prompt embeddings and the CLIP embeddings of input images.

CLIP embedding space includes perceptual linguistic representations, and text-based image-processing tasks are explored. In image quality assessment (IQA), CLIP-IQA \cite{wang2022exploring} compares image embeddings with text embeddings of "a good photo" and "a bad photo" to derive IQA scores. It has achieved competitive performance across various IQA benchmarks. Furthermore, a CLIP-based color control method \cite{Lee_2024_CVPRCLIPTONE} has been proposed that enables color adjustments based on perceptual text expressions. However, the direct use of text prompts for image enhancement remains largely unexplored.  

Another limitation of CLIP-based image enhancement methods \cite{Liang_2023_ICCV,Morawski_2024_CVPR,morawski2024} is that they use architectures such as U-Net or Zero-DCE, which require numerous convolution operations on the image. Consequently, the processing time increases with the image size. To overcome this limitation, we adopt an image-adaptive processing approach that estimates the parameters of simple processing from a small-sized input image. For simple processing, tone curve adjustment is used for image-adaptive approaches \cite{Cui_2024_BMVC,ogino2024erupyolo}. Ogino et al. \cite{ogino2024erupyolo} proposed a B\'{e}zier curve-based tone curve adjustment module for the entire image. This module has demonstrated its effectiveness not only in improving object detection performance but also in data augmentation. However, we observed that applying this process only once to the entire image can result in limited or excessive changes. Our observations indicate that multiple applications of simple processing are necessary to achieve more effective image enhancement.

For this task, we adopt reinforcement learning (RL). It allows for sequential decision-making for the iterative enhancement process. RL has been applied to image processing tasks such as auto exposure \cite{Lee_2024_CVPR}, image signal processing (ISP) \cite{wang2024adaptiveisp}, image editing \cite{hu2018exposure} and image enhancement \cite{zhang2021rellie}. Inspired by these approaches, our method uses RL to iteratively estimate tone curve adjustment parameters (actions) based on the current image state. In our framework, the agent learns a policy to maximize rewards derived from minimizing the distance between the enhanced image and CLIP text embeddings representing 'good' images.

In this paper, we propose CLIP-Utilized Reinforcement learning for Visual image Enhancement (CURVE). CURVE employs a simple image processing module which adjusts global image tone based on Bézier curve and estimates its processing parameters iteratively. The estimator is trained by RL with rewards designed by CLIP text embeddings. Our experiments demonstrate that our method has high enhancement capability for both low-light and multi-exposure datasets. Furthermore, we demonstrate that CURVE has achieved fast processing speed especially on high-resolution images.

\section{Proposed method}
\label{sec:prop}
\begin{figure*}
    \centering
        \begin{tabular}{cc}
             \includegraphics[width=0.6\linewidth]{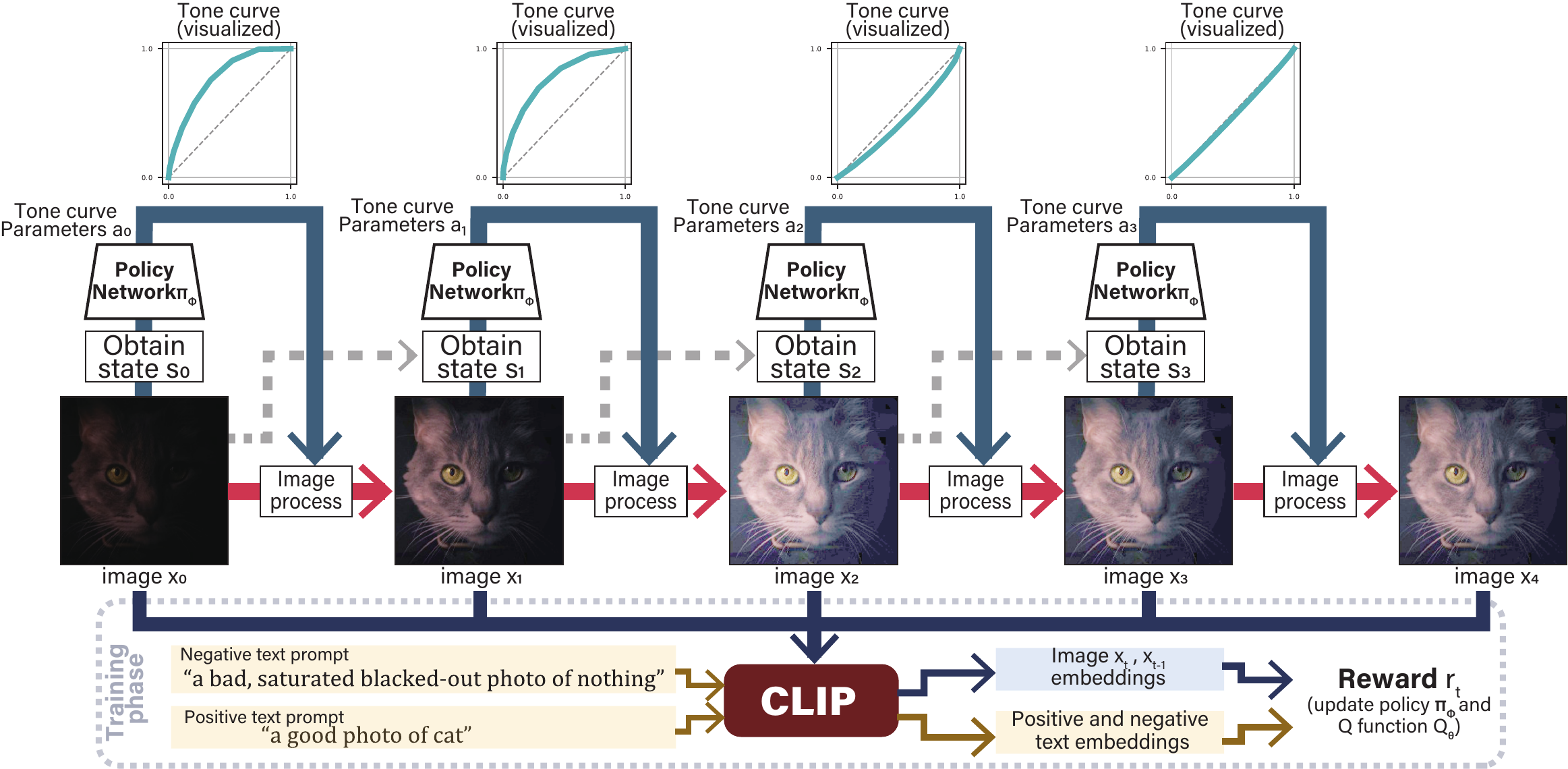}
             &
             \includegraphics[width=0.22\linewidth]{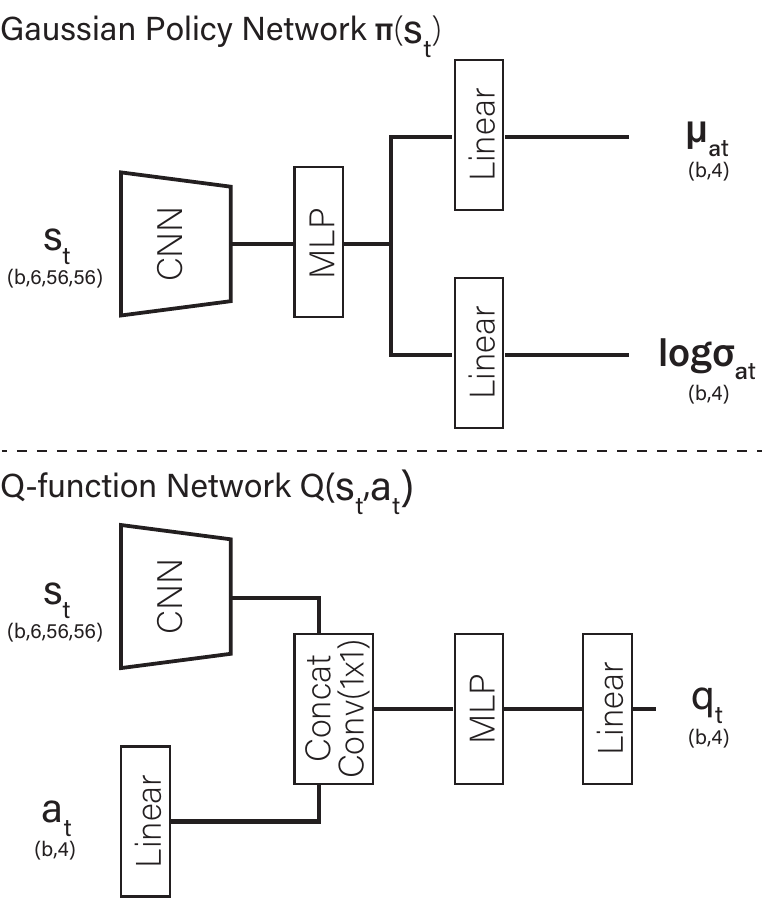}
             \\
             (a)&(b)
        \end{tabular}
    \caption{
    The CURVE method. (a) Framework of CURVE ($T=4$). (The cat image is from ExDark dataset \cite{Exdark}). (b) Architecture of the policy network $\pi_{\phi}$ and Q-function network $Q_{\theta}$. $b$ denotes batch size. We provide details in supplementary materials.}
    \label{fig:framework}
\end{figure*}

Our proposed CURVE employs a reinforcement learning (RL) framework to iteratively apply tone curve adjustments based on B\'{e}zier curves. Fig. \ref{fig:framework}(a) illustrates an overview of our approach. CURVE processes an input image through $T$ steps of tone curve adjustments within an episode. At each time step $t$, our framework obtains the current state ${\mathbf s}_t$ from the current image ${\mathbf X}_t$ and the previous image ${\mathbf X}_{t-1}$. Then, it estimates the tone curve parameters (action value) ${\mathbf a}_t$ from the current state ${\mathbf s}_t$ using the policy network $\pi_{\phi}$. Finally, the estimated tone curve parameters are applied to the image ${\mathbf X}_t$ to obtain the next image ${\mathbf X}_{t+1}$. We employ the Soft Actor-Critic (SAC) algorithm \cite{pmlr-v80-haarnoja18b,haarnoja2018soft} for our RL framework.
SAC is an off-policy RL algorithm that maximizes a weighted sum of reward and entropy . Incorporating the policy entropy improves exploration diversity during training. this effectiveness was demonstrated in the prior auto-exposure work \cite{Lee_2024_CVPR} for continuous action spaces. SAC trains the policy network $\pi_{\phi}$ and the Q-function network $Q_{\theta}$.

\subsection{B\'{e}zier Curve-based Tone Curve Adjustment}
Our B\'{e}zier Curve-based module is adapted from the work of Ogino et al. \cite{ogino2024erupyolo}. This module approximates a B\'{e}zier curve using piecewise linear segments. The module first partitions the parametric space into $L+1$ equally spaced points, defined as ${\mathbf q} = \left\{ \frac{j}{L} \,\middle|\, j = 0,1,\ldots,L \right\}.$
For each point $q_j \in {\mathbf q}$, the corresponding mapping points for pixel input $c_{P_i}$ and pixel output $c_{P_o}$ using cubic B\'{e}zier curve equations are defined as:
\begin{eqnarray}
c_{P_i}(q_j) &=& 3q_j(1-q_j)^2 p^i_1 + 3q_j^2(1-q_j) p^i_2 + q_j^3, \\
c_{P_o}(q_j) &=& 3q_j(1-q_j)^2 p^o_1 + 3q_j^2(1-q_j) p^o_2 + q_j^3
\end{eqnarray}
where $p^i_k$ and $p^o_k (k = 1, 2)$ are the control points for the input and output B\'{e}zier curves, respectively. 
The applied image ${\mathbf X}_t$ using the obtained curve points is expressed as:
\begin{eqnarray}
\label{eq:bezproc}
&{\mathbf X}_{t+1}& =  \nonumber \\
&\Sigma^{L-1}_{j=0}&clip \left({\mathbf X}_t -c_{P_i}(q_j), 0, \Delta c_{P_i}(q_j) \right)\cdot \frac{\Delta c_{P_o}(q_j)}{\Delta c_{P_i}(q_j)},
\end{eqnarray}
where $clip({\mathbf x}, a, b)$ denotes the clipping function that clamps the input ${\mathbf x}$ to the range $[a, b]$ and $\Delta c_{P_m}(q_j)$ denotes the forward difference $c_{P_m}(q_{j+1}) - c_{P_m}(q_j)$.

\subsection{Design for Reinforcement Learning}
\subsubsection{Action and State}
To define the action values ${\mathbf a}_t$ for the B\'{e}zier curve, we adopt parameters $\theta_k$ and $r_k$ ($k = 1, 2$) following prior work \cite{ogino2024erupyolo}. These parameters are set such that the mapping curve becomes the identity mapping when they are zero. We represent these parameters as the action vector ${\mathbf a}_t = [\theta_1, \theta_2, r_1, r_2]$. The control points ${\mathbf p}_k = [p^i_k, p^o_k]$ (k=1,2) of the B\'{e}zier curve are then defined as:
\begin{eqnarray}
\label{eq:points1}
{\mathbf p}_1 &= [\frac{r_1+1}{2} \cos (\frac{(\theta_1+1)\pi}{4}), \frac{r_1+1}{2} \sin (\frac{(\theta_1+1)\pi}{4})], \\
\label{eq:points2}
{\mathbf p}_2 &= [1-\frac{r_2+1}{2} \cos (\frac{(\theta_2+1)\pi}{4}), 1- \frac{r_2+1}{2} \sin (\frac{(\theta_2+1)\pi}{4})].
\end{eqnarray}
We obtain ${\mathbf a}_t$ from the policy network $\pi_{\phi}$, derive the control points ${\mathbf p}_k$ by Eq. \ref{eq:points1}, \ref{eq:points2}, and process the image by Eq. \ref{eq:bezproc}.

We define the image state ${\mathbf s}_t$ using ${\mathbf X}_t$ and ${\mathbf X}_{t-1}$.
Let ${\mathbf x}_t \in \mathbb{R}^{C \times h \times w}$ be the resized image of ${\mathbf X}_t \in \mathbb{R}^{C \times H \times W}$. We define ${\mathbf v}_t = {\mathbf x}_t - {\mathbf x}_{t-1}$, with ${\mathbf v}_0 = \mathbf{0} \in \mathbb{R}^{C \times h \times w}$. The state ${\mathbf s}_t \in \mathbb{R}^{2C \times h \times w}$ is a tensor formed by concatenating ${\mathbf x}_t$ and ${\mathbf v}_t$ along the channel dimension.

\subsubsection{CLIP-based Reward}

We design a novel reward function inspired by CLIP-IQA \cite{wang2022exploring}, using both negative and positive text prompts. The reward is designed to increase as the image becomes closer to the positive text prompts in the CLIP embedding space. Additionally, we utilize an object detection dataset for training. We incorporate object names from the dataset into the positive prompts to make them more specific to the input image content.

The negative text prompt $T_n$ is fixed as "a bad, saturated, blacked out photo of nothing". For each object class $i$ included in the input image, we define a positive text prompt $T_{p_i}$ as "a good photo of \{class $i$ name\}". Let $N$ be the number of unique object classes detected in the input image ${\mathbf X}_t$ at time step $t$. 

We define the loss $L_t$ at time step $t$ as:
\begin{equation}
    \label{eq:loss}
    L_t = \frac{1}{N}\sum_{i=1}^N - \log \left(
    \frac{e^{m(f_{T_{p_i}},f_{{\mathbf X}_t})}}{\sum_{v \in \{p_i,n\}}e^{m(f_{T_v},f_{{\mathbf X}_t})}}
    \right),
\end{equation}
where $m(\cdot,\cdot)$ denotes the cosine similarity, $f_{T_{p_i}}$ is the CLIP text embedding of the positive prompt for class $i$, $f_{{\mathbf X}_t}$ is the CLIP image embedding of ${\mathbf X}_t$, and $f_{T_n}$ is the CLIP text embedding of the negative prompt.
This loss function can be interpreted as the cross-entropy loss applied to the softmax probabilities of the cosine similarities between the image embedding  $f_{{\mathbf X}_t}$ and the text embeddings $f_{T_{p_i}}$ and $f_{T_n}$.

The reward $r_t$ is calculated as the improvement in the loss:
\begin{equation}
    r_t = \beta (L_{t} - L_{t+1}),
\end{equation}
where $\beta$ is a scaling factor (set to $200$ in our experiment).

\subsubsection{Policy and Value Networks}
Fig. \ref{fig:framework}(b) shows the policy and the Q-function networks.
The Q-function network $Q_{\theta}$ (Fig. \ref{fig:framework}(b) bottom)  predicts the expected discounted sum of future rewards plus policy entropy and is used only during training. It outputs a soft state-action value $q_t$ from inputs ${\mathbf s}_t$ and ${\mathbf a}_t$. To integrate the parameter vector ${\mathbf a}_t$ with the two-dimensional image state ${\mathbf s}_t$, we design a network that employs a CNN for ${\mathbf s}_t$ and a linear projection for ${\mathbf a}_t$. The outputs are then concatenated along the channel dimension, followed by a $1 \times 1$ convolution to obtain a one-dimensional vector. This vector is input to a two-layer MLP and a linear projection to obtain $q_t$.

The Q-function network $Q_{\theta}$ is trained to minimize the soft Bellman residual \cite{haarnoja2018soft}:
\begin{eqnarray}
   &J_Q(\theta) = {\mathbb E}_{({\mathbf s}_t,{\mathbf a}_t)\sim {\mathcal D}}& \Bigl[ 
   \frac{1}{2}
   \Bigl(
   Q_{\theta}({\mathbf s}_t,{\mathbf a}_t)- \nonumber\\
   &&\bigl( r_t + 
   \gamma 
   {\mathbb E}_{s+1\sim p}\left[V_{\overline{\theta}}({\mathbf s}_{t+1})\right] \bigr) \Bigr)^2 
   \Bigr],
\end{eqnarray}
\begin{eqnarray}
    &{\mathbb E}_{s+1\sim p}& \left[V_{\overline{\theta}}({\mathbf s}_{t+1}) \right] = \nonumber \\ 
    &&Q_{\overline{\theta}}({\mathbf s}_{t+1},{\mathbf a}_{t+1}) - \alpha \log \left( \pi_{\phi}({\mathbf a}_{t+1}|{\mathbf s}_{t+1}) \right), 
\end{eqnarray}
where $\gamma$ is the discount rate, and $Q_{\overline{\theta}}$ is a target soft Q-function network with parameters $\overline{\theta}$ that are obtained as an exponentially moving average of the soft Q-function weight $\theta$.

The policy network $\pi_{\phi}$ illustrated in the top of Fig. \ref{fig:framework}(b) outputs ${\mathbf a}_t$ from ${\mathbf s}_t$. The network first obtains a feature vector from ${\mathbf s}_t$ through a CNN, and then inputs it to a two-layer MLP. The features are used to obtain ${\mathbf \mu}_t$ and $\log {\mathbf \sigma}_t$ through separate linear projections. The policy network $\pi_{\phi}$ follows the Gaussian policy\cite{haarnoja2018soft}, sampling actions from $\mathcal{N}({\mathbf \mu}_t, {\mathbf \sigma}_t)$ during training and using ${\mathbf \mu}_t$ at test time. The action value ${\mathbf a}_t$ is obtained by applying $tanh$ and rescaling the output.

\begin{figure}[t]
    \begin{algorithm}[H]
        \caption{Test Time Algorithm}
        \label{alg1}
        \begin{algorithmic}[1]   
        \REQUIRE image ${\bf X}_0$, \\
        initial state ${\bf s}_0 = concat({\bf x}_0,{\bf 0})$, LUT vector ${\bf l}_0$ \\
        \FOR{$t=0;t<T;t++$}
        \STATE ${\bf a}_t \leftarrow \pi_{\theta}({\bf s}_t)$
        \STATE ${\bf x}_{t+1} \leftarrow B({\bf x}_t, {\bf a}_t)$
        \STATE ${\bf v}_{t+1} \leftarrow {\bf x}_{t+1}-{\bf x}_{t}$
        \STATE ${\bf s}_{t+1} \leftarrow concat({\bf x}_{t+1}, {\bf v}_{t+1})$ 
        \STATE ${\bf l}_{t+1} \leftarrow B({\bf l}_t, {\bf a}_t)$
        \ENDFOR
        \STATE ${\bf X}_T \leftarrow Map({\bf X}_0, {\bf l}_T)$
        \end{algorithmic}
    \end{algorithm}
\end{figure}

The policy network $\pi_{\phi}$ is trained to minimize the following function \cite{haarnoja2018soft} so that it selects actions with both high Q-values and high entropy:
\begin{equation}
\begin{split}
J_{\pi}(\phi) = 
{\mathbb E}_{{\mathbf s}_t \sim {\mathcal D}} 
\left[
{\mathbb E}_{{\mathbf a}_t \sim \pi_{\phi}}
\left[
\alpha \log \left( \pi_{\phi} ({\mathbf a}_t|{\mathbf s}_t) \right) - Q_{\theta}({\mathbf s}_t, {\mathbf a}_t)
\right]
\right].
\end{split}
\end{equation}
Here, $\alpha$ is a trainable parameter.

\begin{table*}[t]
    \centering
    \small
    \caption{Comparison of CURVE with conventional LLIE methods on low-light datasets. The left side of the table shows paired results and the right side show processing times per frame for HD ($720\times1280$), FHD ($1080\times1920$), and UHD ($2160\times3840$) resolutions on GPU. Note that ReLLIE encountered memory errors for UHD resolution. The 'train-by-loss' method has the same processing time as the proposed method. The best and the second scores are bold and underlined, respectively.}
    \label{tb:res_lol}
    \begin{tabular}{ll|cc|cc|cc|ccc}
    &&\multicolumn{2}{c|}{lolv1\cite{Chen2018Retinex}}&\multicolumn{2}{c|}{lolv2real\cite{9328179}}&\multicolumn{2}{c|}{lolv2syn\cite{9328179}}&\multicolumn{3}{c}{Runtime [sec/frame]}\\ 
    &&{SSIM}&{PSNR}&{SSIM}&{PSNR}&{SSIM}&{PSNR}&HD&FHD&UHD
    \\
    \hline
    \multirow{3}{*}{Zero reference
    }
    &{Zero-DCE \cite{Guo_2020_CVPR}}&0.6644&14.86&\underline{0.6768}&\underline{18.06}&\underline{0.8399}&\underline{17.76}&
    0.020&0.045&0.205\\
    &{RUAS \cite{Liu_2021_CVPR}}&\underline{0.7035}&16.40&0.6727&15.33&0.6763&13.40&
    0.021&0.045&0.184\\
    &{SCI \cite{Ma_2022_CVPR}}&0.6166&14.78&0.6286&17.30&0.7658&15.43&
    {\bf 0.002}&{\bf 0.005}&\underline{0.020}\\
    \multirow{2}{*}{CLIP-based
    }&{CLIP-LIT \cite{Liang_2023_ICCV}}&0.5884&12.39&0.6334&15.18&0.7965&16.19&
    0.064&0.144&20.68\\
    &{Morawski \cite{Morawski_2024_CVPR}}&0.6265&13.48&0.6545&17.19&0.8045&17.43&
    0.020&0.044&0.193\\
    {RL-based}&
    {ReLLIE \cite{zhang2021rellie}}&0.6746&{\bf 18.74}&0.6365&16.79&0.8360&15.64&1.436&3.303&-\\
    {Ablation study}&{train-by-loss}&0.5870&12.56&0.6173&15.84&0.7819&16.99&-&-&-\\
    &{CURVE (ours)}&{\bf 0.7164}&\underline{17.35}&{\bf 0.6794}&{\bf 18.83}&{\bf 0.8487}&{\bf 17.92}&
    \underline{0.016}&\underline{0.017}&{\bf 0.017}\\
    \hline
    \end{tabular}
\end{table*}
\begin{figure*}
    \centering
        \begin{tabular}{cc}
        \includegraphics[width=0.4\linewidth]{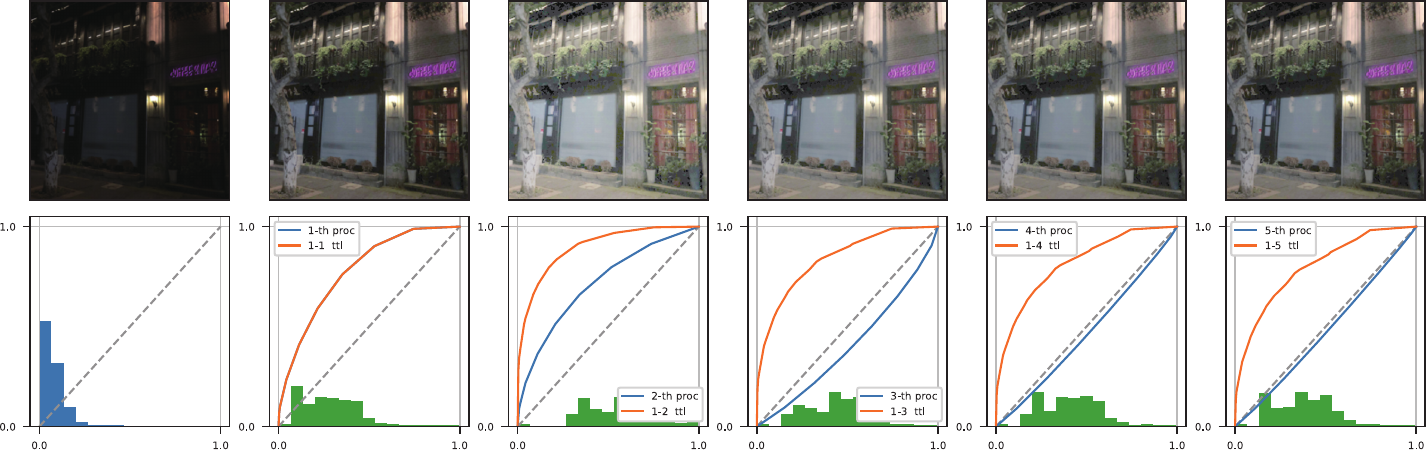}&
        \includegraphics[width=0.4\linewidth]{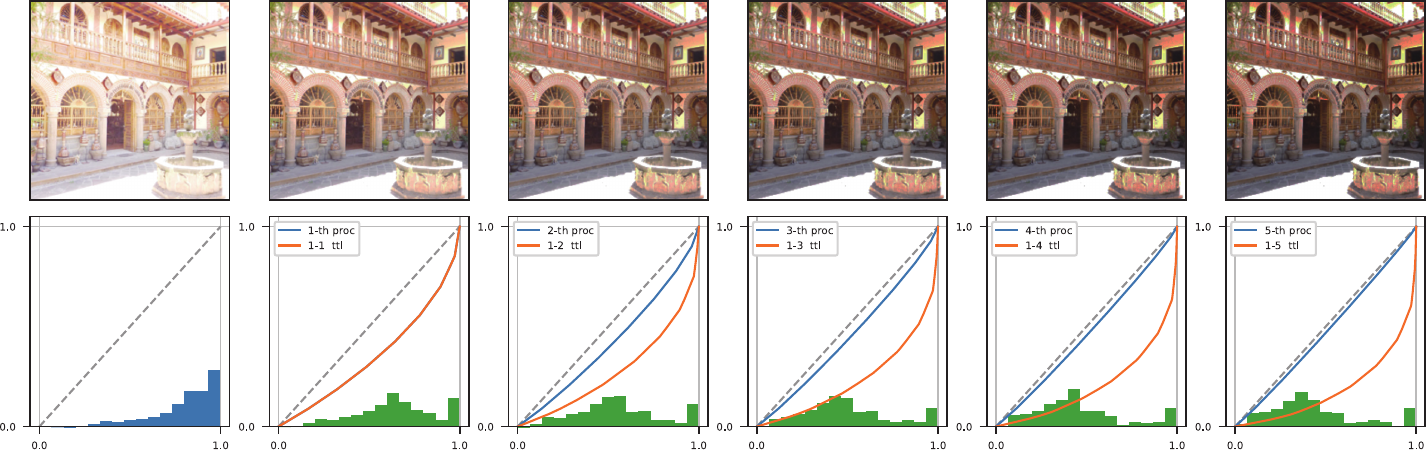}\\
        (a)&(b)
        \end{tabular}
    \caption{Application of CURVE to (a) under-exposed and (b) over-exposed images. The top row shows the images during the enhancement process. The bottom row displays the applied tone curves (blue lines), cumulative curves relative to $t=0$ (orange lines), and image histograms at each step.}
    \label{fig:toneexamples}
\end{figure*}

\subsection{Implementation for Fast Test-Time Processing}
Our tone adjustment processing modifies the mapping function of image pixel values. This property enables efficient computation at test time. It can derive the next state ${\mathbf s}_{t+1}$ directly from the current small-size image ${\mathbf x}_t$ and the processing parameters ${\mathbf a}_t$, without processing the full-resolution image ${\mathbf X}_t$. Furthermore, it enables the computation of a composed Look-Up Table (LUT) through every time step for all integer pixel values. We define the initial LUT vector as ${\mathbf l}_0 = \left\{ {i}/{(2^{bit}-1)} \,\middle|\, i = 0,1,\ldots,2^{bit}-1 \right\}$ where $bit$ is the bit depth of the image. By applying the sequence of actions ${\mathbf a}_t$ to this vector, we obtain the composed ${\mathbf l}_t$ at each time step. This process allows the construction of a composite LUT from time step $0$ to $T$.

Algorithm \ref{alg1} outlines the processing method. Here, $B({\mathbf x}_t, {\mathbf a}_t)$ denotes the function of tone curve adjustment with parameters ${\mathbf a}_t$ to ${\mathbf x}_t$, and $Map({\mathbf X}_0, {\mathbf l}_T)$ represents the replacement of luminance values in the original image ${\mathbf X}_0$ using the final lookup table ${\mathbf l}_T$.

\def\ims{0.31\linewidth}

\begin{figure*}
\centering
\begin{tabular}
{m{0.03\linewidth}m{\ims}m{\ims}}
\centering
    {\scriptsize Original}
    &
     \includegraphics[width=\linewidth]{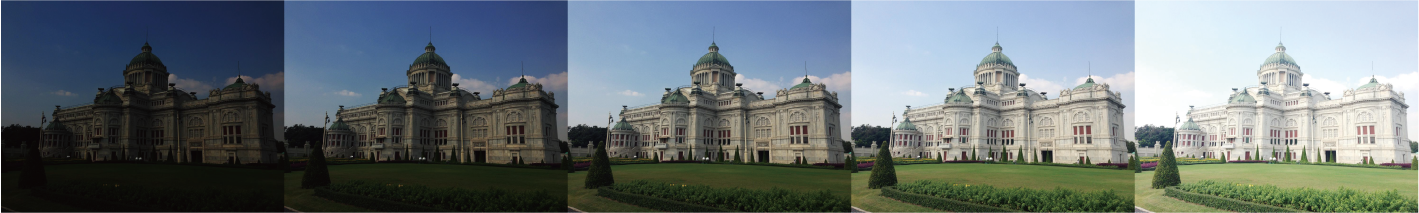}&
     \includegraphics[width=\linewidth]{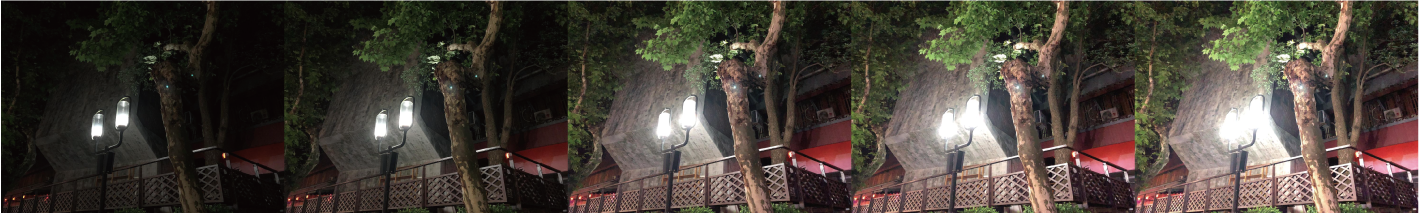}
     \\
     {\scriptsize Zero-DCE\cite{Guo_2020_CVPR}}
     &
     \includegraphics[width=\linewidth]{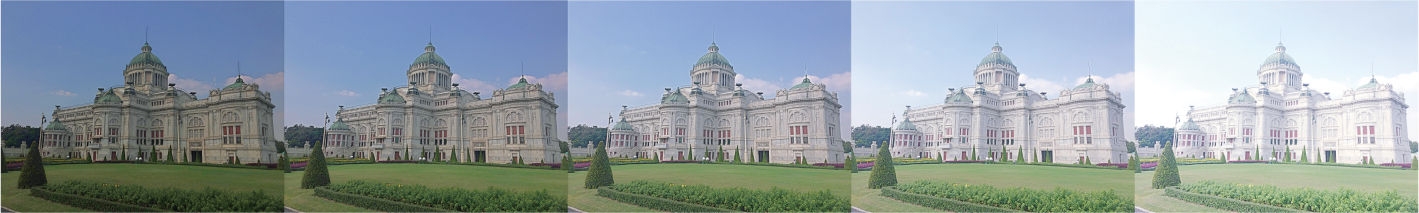}&
     \includegraphics[width=\linewidth]{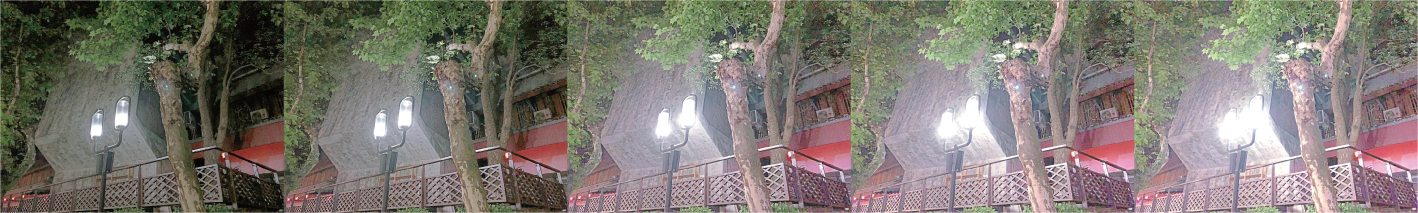}
     \\ 
     {\scriptsize CURVE}
     &
     \includegraphics[width=\linewidth]{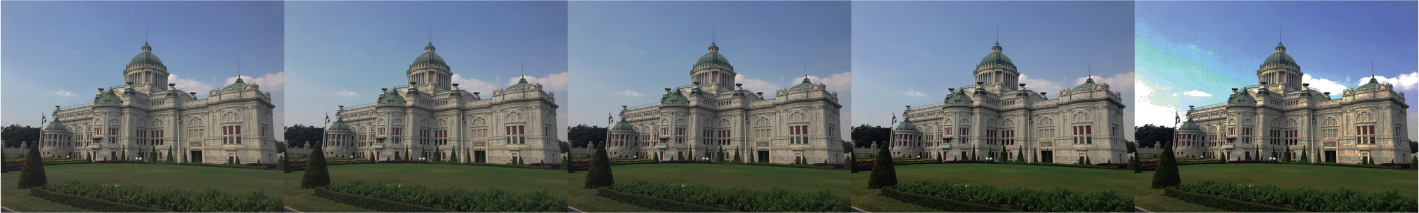}&  
     \includegraphics[width=\linewidth]{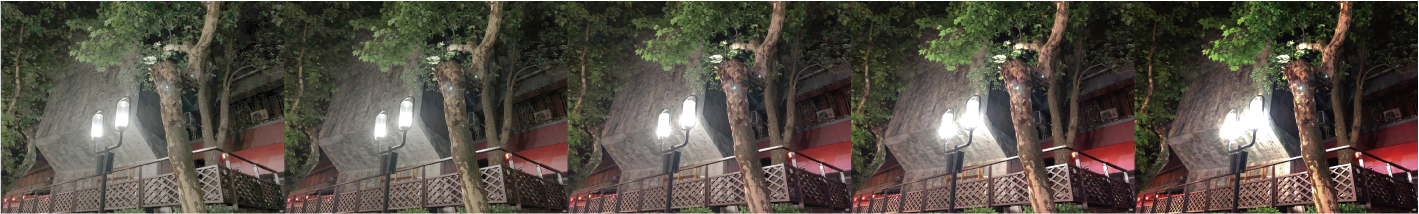}\\
\end{tabular}
\caption{Visual results on multi-exposure images from the SICE dataset \cite{8259342}. Other results are available at supplementary materials.}
\label{fig:views}
\end{figure*}

\section{Experiments}

\subsection{Experimental Settings}
{\bf Implementations and training details.} We used the VOC 2007 trainval dataset \cite{vocdataset} for training. Following the training method in prior work \cite{ogino2024erupyolo}, we employed the tone curve module for data augmentation. We applied random parameters by sampling from a normal distribution ${\mathcal N}(0,1)$. The hidden layers of MLPs and the output feature size of CNNs were set to 256. The architecture of CNN is described in Table \ref{tb:cnn}. We employed the RN50 CLIP model for feature extraction. During training, input images ${\mathbf X}_t$ were center-cropped to a size of $(3,224,224)$. The downsample size of ${\mathbf x}_t$ was set to $(3,56,56)$. We trained the policy network $\pi_{\phi}$ and Q-function network $Q_{\theta}$ using the SAC algorithm with the hyperparameters listed in Table \ref{tb:settings}. For evaluation, test images were directly center-cropped to a size of $(3,56,56)$ to obtain ${\mathbf x}_t$.
\begin{table}[t]
\centering
\small
\caption{Training parameters of SAC\cite{haarnoja2018soft}}
\label{tb:settings}
\begin{tabular}{lr}
Parameter&Value\\
\hline
optimizer&Adam\\
learning rate&$3 \cdot 10^{-4}$\\
discount rate $\gamma$&0.99\\
replay buffer size&$10^6$\\
batch size $b$ &$256$\\
target smoothing coefficient $\tau$& $5\cdot 10^{-4}$\\
target update interval/gradient steps&1\\
episode steps $T$&5\\
$a_t$ value range&$[-2,2]$\\
total iterations&$7.5 \cdot 10^{5}$\\
policy training start steps & $10^4$\\
\hline
\end{tabular}
\end{table}
\begin{table}[t]
\footnotesize
\centering
\caption{CNN Architecture. ReLU between each layer.}
\label{tb:cnn}
\begin{tabular}{lcccc}
&{  in-ch}&{  out-ch}&{  ksize}&{  stride}\\
\hline
{Conv2d}&6&8&7&3\\
{Conv2d}&8&16&3&2\\
{Conv2d}&16&16&3&1\\
{Conv2d}&16&32&3&1\\
{Conv2d}&32&256&3&1\\
\multicolumn{5}{c}{  Adaptive Pool2d}\\
{  linear}&256&256&&\\
\hline
\end{tabular}
\end{table}

{\bf Evaluations.} We evaluated our method on multiple datasets: the test sets of LOLv1 \cite{Chen2018Retinex}, LOLv2Real \cite{9328179}, and LOLv2Syn \cite{9328179} for low-light enhancement, and the SICE Part.2 dataset \cite{8259342} for multi-exposure image enhancement. We compared our approach against several zero-reference LLIE techniques: Zero-DCE \cite{Guo_2020_CVPR}, RUAS \cite{Liu_2021_CVPR}, SCI \cite{Ma_2022_CVPR}, CLIP-LIT \cite{Liang_2023_ICCV}, CLIP-based LLIE by Morawski et al. \cite{morawski2024} and ReLLIE \cite{zhang2021rellie} (Zero-DCE incorporated into an RL framework). We used the pre-trained models provided for the low-light enhancement task for all these methods. For the multi-exposure evaluation using the SICE dataset, we compared our method only with Zero-DCE, as it is trained on this dataset and capable of handling multi-exposure images. Following the protocol of Zero-DCE \cite{Guo_2020_CVPR}, we resized the images of SICE to (900, 1200).
To validate the effectiveness of our approach, we included a 'train-by-loss' model in our comparison, which optimizes $\pi_{\phi}$ using the loss function defined in Equation \ref{eq:loss}. The training parameters were almost the same in Table \ref{tb:settings}, but the number of iterations was 15 k. 

To assess computational speed, we compared processing times for major display resolutions: HD, FHD, and UHD. Timing starts from loading an image array onto GPU memory and ends when the processed uint8 arrays were retrieved on GPU. As an exception, for ReLLIE \cite{zhang2021rellie}, we measured the time from loading images on CPU to outputting them on CPU. This is due to its hard-coded CPU/GPU allocation in the RL framework. Our evaluation environment comprises an Intel Core i9-9900K CPU @3.60GHz and an NVIDIA GeForce RTX 3080 GPU. The processing time per frame was calculated as the average processing time of 100 times.

\subsection{Experimental Results}
\label{sec:exp}
Table \ref{tb:res_lol} presents the SSIM and PSNR values\footnote{we used  \url{https://github.com/chaofengc/IQA-PyTorch}.} for each dataset. Our proposed method achieves either the best or second-best performance across all datasets. As shown in the right side of Table \ref{tb:res_lol}, our approach is faster than most competing methods. The results on the SICE dataset, presented in Table \ref{tb:res_sice}, further validate the effectiveness of our method in handling multi-exposure scenarios. Our approach is outperformed by the baseline methods. Fig. \ref{fig:views} shows the visual results on the SICE dataset. Our method demonstrates the ability to maintain consistent brightness across different exposure levels. Fig. \ref{fig:toneexamples} provides a step-by-step visualization of processing for under-exposed and over-exposed images. These visualizations demonstrate that our method succeeds in adaptive processing on the input image.

\begin{figure}
    \centering
        \includegraphics[width=0.75\linewidth]{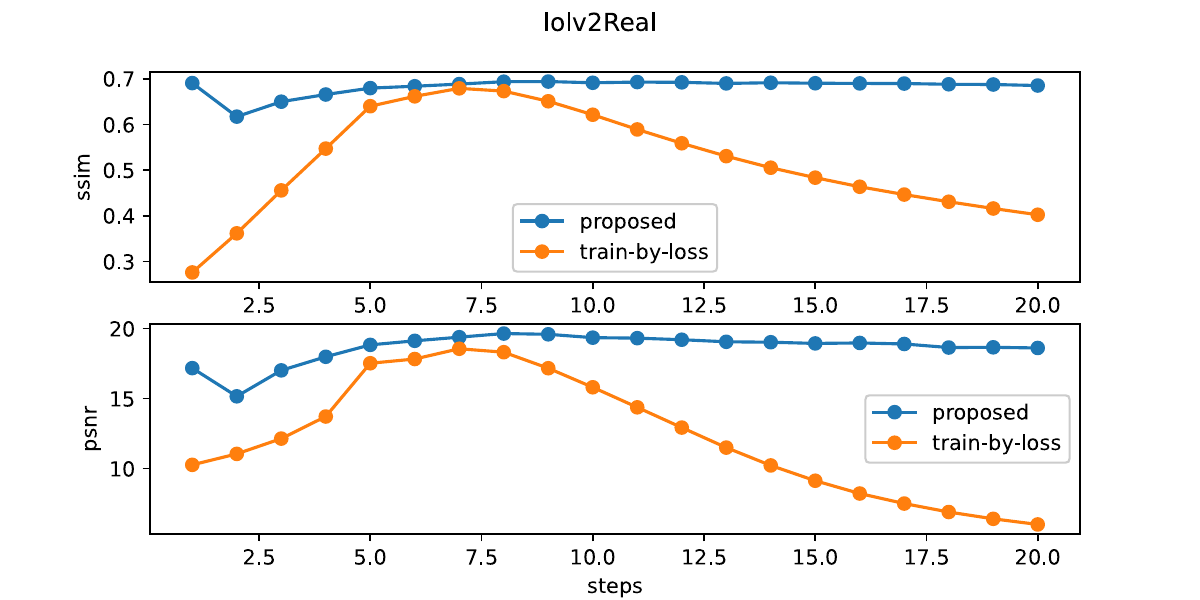}
    \caption{
    SSIM and PSNR at multiple time steps.
    }
    \label{fig:ablation}
\end{figure}

To further validate our RL approach, we compare our method with a "train-by-loss" model. Fig. \ref{fig:ablation} shows the progression of PSNR and SSIM values for both models over 20 steps. Our proposed method (blue line) converges quickly to high-performance values and maintains stability even with increasing steps. In contrast, the "train-by-loss" model (orange line) shows a gradual improvement followed by degradation after reaching a peak, indicating potential overfitting or instability in extended processing.

\begin{table}[t]
\centering
\footnotesize
\caption{SICE\cite{8259342} multi-exposure dataset results.}
\label{tb:res_sice}
\begin{tabular}{lccc}
&{SSIM$\uparrow$}&{PSNR$\uparrow$}&{LPIPS$\downarrow$}\\
\hline
{Zero-DCE\cite{Guo_2020_CVPR}}&\underline{0.5615}&\underline{14.02}&\underline{0.2967}\\
{train-by-loss}&0.5283&12.46&0.3460\\
{CURVE(ours)}&{\bf 0.5684}&{\bf 15.81}&{\bf 0.2737}\\
\hline
\end{tabular}
\end{table}

\section{Conclusion}
\label{sec:conc}
In this paper, we introduced CURVE, a novel image enhancement method combining reinforcement learning with CLIP-based reward design. Our approach addresses both low-light and multi-exposure image enhancement challenges while maintaining computational efficiency. CURVE utilizes a fast, adaptive processing framework with B\'{e}zier curve-based tone curve adjustments and a CLIP-guided reward system. Experimental results demonstrate competitive or superior performance compared to state-of-the-art methods across various datasets, with significant improvements in processing speed for high-resolution images. However, as shown in the over-exposed images in Fig. \ref{fig:views}, our method has limitations due to its global processing approach. It may amplify noise or tonal differences in the image. Despite these limitations, CURVE has achieved high-quality enhancement.

\bibliographystyle{ieee}
\bibliography{reference}

\setcounter{figure}{4}
\setcounter{table}{4}
\setcounter{equation}{10}
\setcounter{section}{4}

\makesuptitle
This supplementary material provides additional details and experimental results for our CURVE method. We first explain details of our tone curve module and reinforcement learning framework that were omitted from the main paper. Then, we demonstrate visual comparisons for both multi-exposure and low-light datasets..

\section{Supplementary Details}
\subsection{Bézier-Curve Tone Adjustment (Sec. 2.1)}
Tone adjustment maps the intensity of an input image to output image values, thereby changing the image contrast. Our method uses a cubic Bézier curve to define this mapping function. A cubic Bézier curve is a parametric curve whose shape can be modified by adjusting the positions of its control points. For a parameter $q$ ($0\leq q \leq1$), the two-dimensional curve coordinates $c_P(q)=[x(q),y(q)]$ of a cubic Bézier curve can be expressed using four control points $p_i$ as follows:
\begin{eqnarray}
c_P(q) =&  \nonumber\\
(1-q)^3 p_0& + 3q(1-q)^2 p_1 + 3q^2(1-q) p_2 + q^3 p_3
\end{eqnarray}
Since $p_0$ and $p_3$ represent the start and end points of the curve, we set $p_0=[0,0]$ and $p_3=[1,1]$ to fix the tone curve’s endpoints at the origin and unit point. This yields Eqs. 1 and 2 in the main paper. Our tone adjustment module changes the values of $p_1$ and $p_2$ to modify the shape of the curve, thereby altering the mapping function. Fig. \ref{fig:tone_ex} shows examples of applying different mapping functions to an image. As illustrated, the contrast changes according to the curve shape.

\begin{figure}[t]
\centering
    \includegraphics[width=0.7\linewidth]{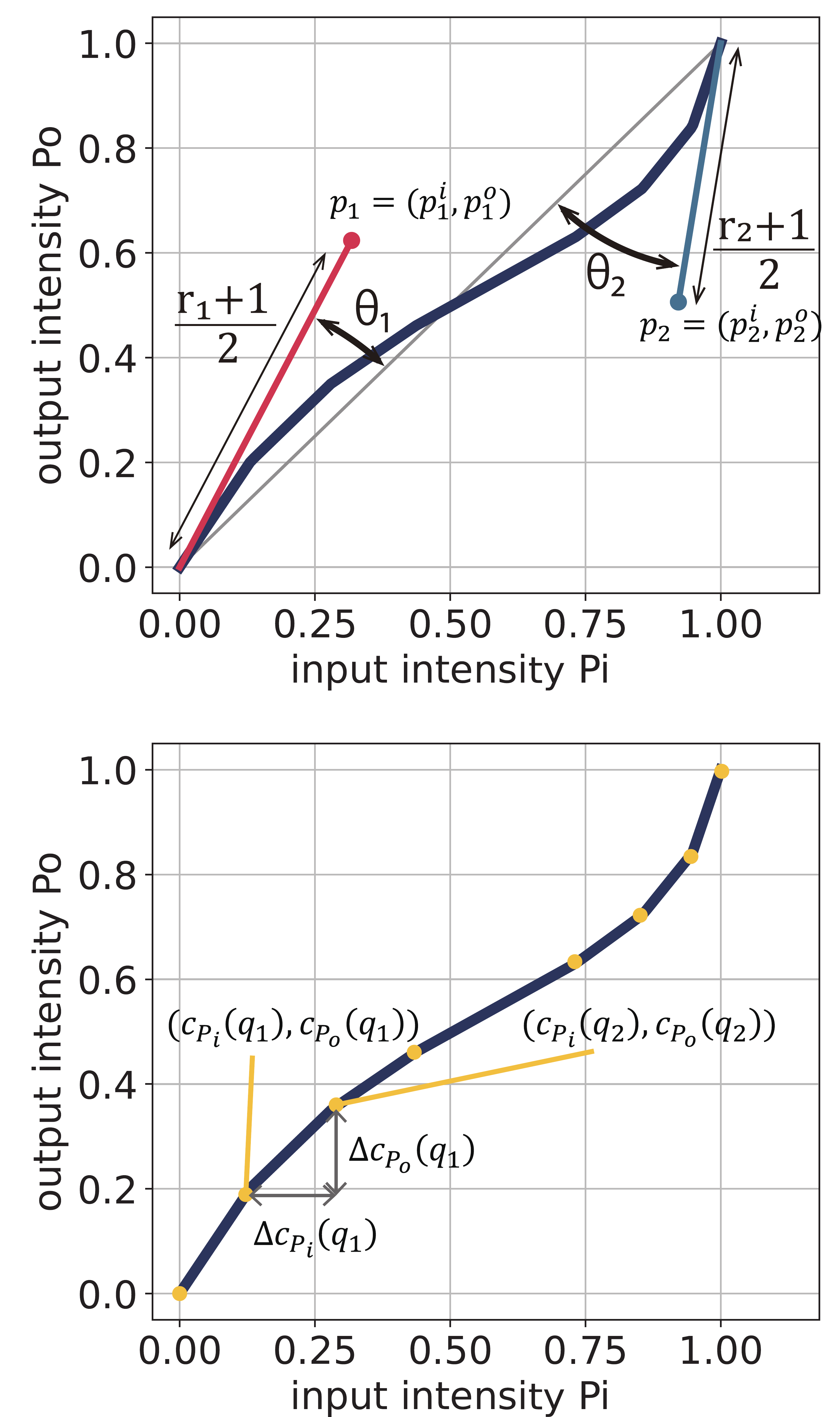}
    \caption{Definition of our Bézier curve tone adjustment. The upper figure shows parameterization using $r_i$ and $\theta_i$ for control points $p_1$ and $p_2$. The lower part illustrates the relationship between input and output intensity points $cP_i(q_j)$ and $cP_o(q_j)$ (as defined in Eq. 3 of the main paper).
    }
    \label{fig:bezier_def}
\end{figure}

\begin{figure*}
\centering
    \includegraphics[width=0.9\linewidth]{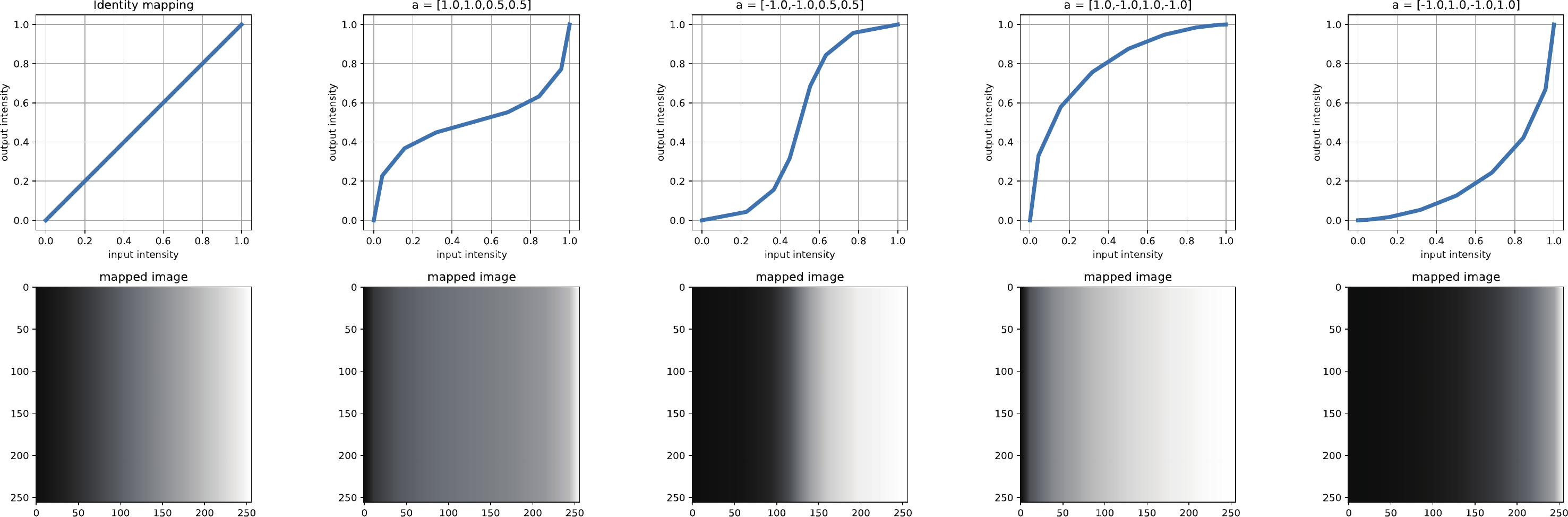}
\caption{Visualization of different tone curves and their effects on a gradient image. The top row shows various tone mapping functions. The leftmost is the identity mapping. The bottom row shows the corresponding output images when these mappings are applied to a gradient image (leftmost bottom). }
    \label{fig:tone_ex}
\end{figure*}

For implementation, our tone adjustment module approximates this Bézier curve using piecewise linear segments. Fig. \ref{fig:bezier_def} illustrates the curve when $L=8$  (seven linear segments, eight sample points). As shown in the lower part of Fig. \ref{fig:bezier_def}, each point of the piecewise linear approximation corresponds to $(c_{P_i}(q_j),c_{P_o}(q_j))$ (as in Eq. 3 of the main paper).

We design the action parameters for reinforcement learning using $r_i$ and $\theta_i$ as illustrated in the upper part of Fig. \ref{fig:bezier_def} (corresponding to Eq. 4 and 5 in the main paper). The purpose of this design is to make the curve shape approach an identity mapping when these values are close to zero.

\subsection{Reinforcement Learning Framework (Sec.2.2)}

We adopted the Soft Actor-Critic (SAC) algorithm as our reinforcement learning (RL) approach, following prior research. Since we made no significant modifications to SAC itself, we kept the algorithm’s description brief in the main paper. This section provides additional details of our RL framework for clarity. In particular, we explain how the CLIP module and the tone adjustment are integrated into the training process, and we describe the inference (testing) framework as well.

RL aims to maximize the expected cumulative reward obtained during an episode consisting of repeated actions and state transitions. Various algorithms have been proposed for this purpose. Soft Actor-Critic (SAC) is an off-policy method that maximizes the expected cumulative reward and policy entropy (diversity of action selection). During training, SAC simultaneously updates two networks: the policy network $\pi_\phi$ and the Q-function network $Q_\theta$ (only $\pi_\phi$ is used during inference). The Q-function network estimates the expected sum of future discounted rewards and policy entropy (the so-called soft Q-value) for a given state ${\bf s}_t$ and action ${\bf a}_t$ (Fig. 1(b) bottom in the main paper). The policy network takes a state as input and outputs parameters $(\mu, \sigma)$ that define a Gaussian distribution over actions. During training, actions are sampled from this distribution for exploration, while during inference (testing), the mean $\mu$ is used for deterministic action selection (Fig. 1(b) top in the main paper).

We integrate CLIP and the tone-adjustment module into the SAC framework shown in Fig. \ref{fig:train_loss}(a). Since SAC is an off-policy method, we utilize past exploration data obtained during training. We store tuples of $[{\bf s}_t, {\bf a}_t,  r_t, {\bf s}_{t+1}, {\bf a}_{t+1}]$ and sample batches from this experience replay buffer when updating each network. CLIP is used as a fixed encoder to evaluate perceptual quality, and no gradients are propagated through the CLIP model.

Fig. \ref{fig:train_loss}(b) illustrates the flow for calculating the loss $L_t$ used for reward design (Sec. 2.2.2 in the main paper). From positive/negative text $T$, we obtain feature vector $f_T$ using CLIP's text encoder, and from image ${\bf X}$, we also obtain image feature vector $f_{\bf X}$ from the CLIP’s image encoder. The cosine similarity $m(\cdot,\cdot)$ between the features is expressed as:
\begin{eqnarray}
m(f_T, f_{\bf X}) = \frac{f_T^T f_{\bf X}}{||f_T|| ||f_{\bf X}||}.
\end{eqnarray}

This similarity is used in a softmax-based loss (see Eq. (6) in the main paper) that encourages the enhanced image feature $f_{\bf X}$ to be closer to the ‘good image’ text feature $f_{T_p}$ than to the ‘bad image’ prompt $f_{T_n}$.

Finally, Fig. \ref{fig:testing} shows the flow during inference (testing). Our method iteratively optimizes the mapping function and the intensity mapping. Our goal is to obtain the final high-resolution image ${\bf X}_T$ with all mapping processes applied, but applying the mapping process to every image ${\bf X}_t$ would be time-consuming. It can create state ${\bf s}_{t+1}$ by applying the mapping function to the downsized image state ${\bf s}_t$ without generating intermediate high-resolution image ${\bf X}_t$. Meanwhile, even when we have obtained ${\bf a}_t$ for all $t$, it is theoretically hard to derive a composite mapping function through an episode (from $t=0$ to $t=T$). Therefore, we take advantage of the discrete value and limited range of pixel values by applying our mappings to a lookup table (LUT) that covers all possible input values. By sequentially applying actions ${\bf a}_t$ to this LUT ${\bf l}$, we create a composite LUT ${\bf l}_T$ from the initial LUT ${\bf l}_0$, which is then applied to the original image ${\bf X}_0$ to obtain the final enhanced result (Sec. 2.2.3 in the main paper).

\section{Extended Qualitative Comparisons}
\subsection{Multi-Exposure Image Dataset}
We show the comparison results on the SICE Part 2 dataset in Fig. \ref{fig:sice-1} and Fig. \ref{fig:sice-2}. While our proposed method consistently maintains equivalent brightness levels across images with different exposures, Zero-DCE tends to further brighten already over-exposed images.

\subsection{Low-Light Image Dataset}
We show the comparison results on the LOLv2 dataset in Fig. \ref{fig:lolv2-1} and Fig. \ref{fig:lolv2-2}. Compared to other low-light image enhancement methods, our proposed approach achieves a brightness balance closer to the ground truth.

\def\ims{0.35\linewidth}

\begin{figure*}
\centering
\begin{tabular}{cc}
    \includegraphics[height=\ims]{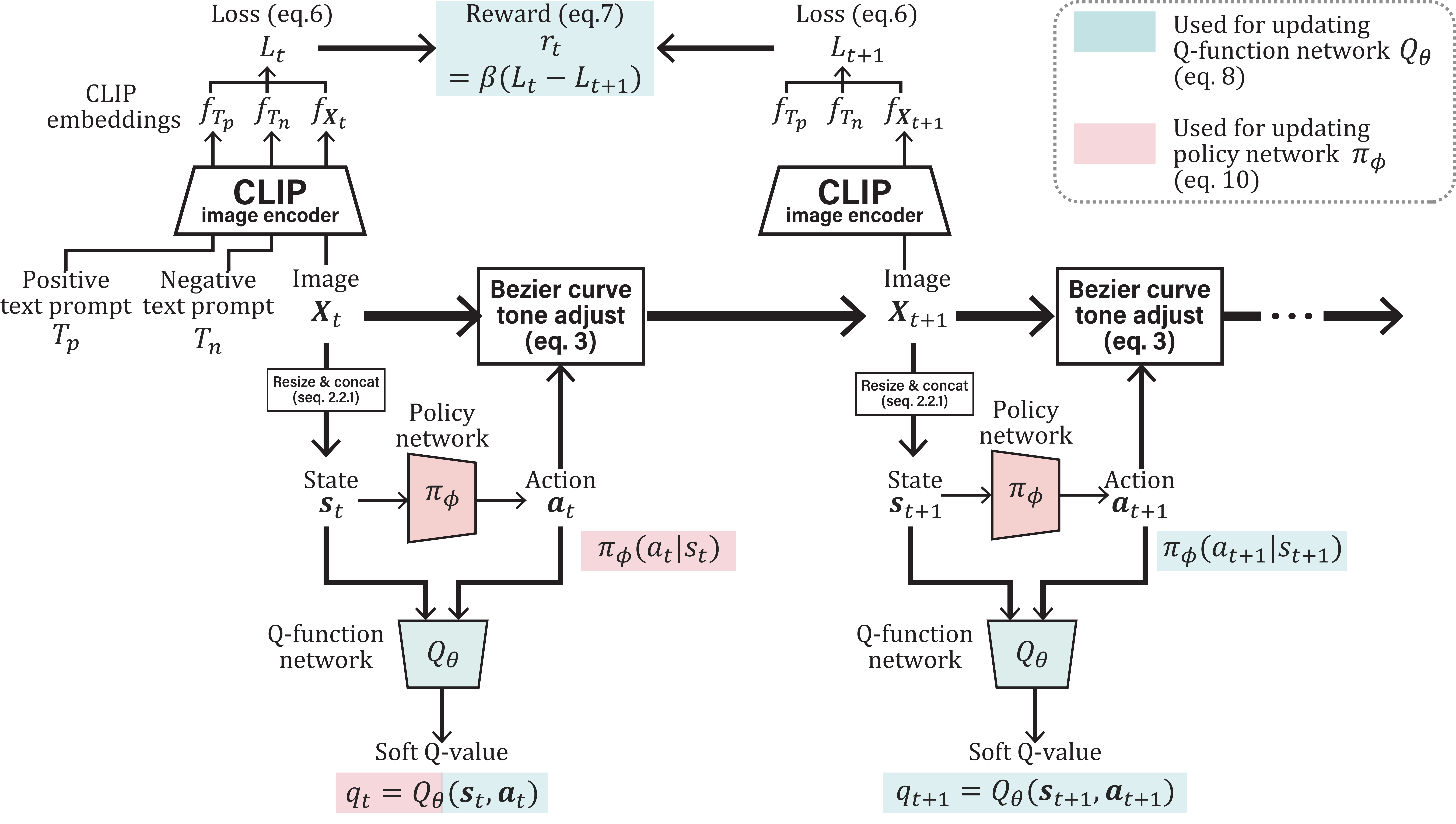}&
    \includegraphics[height=\ims]{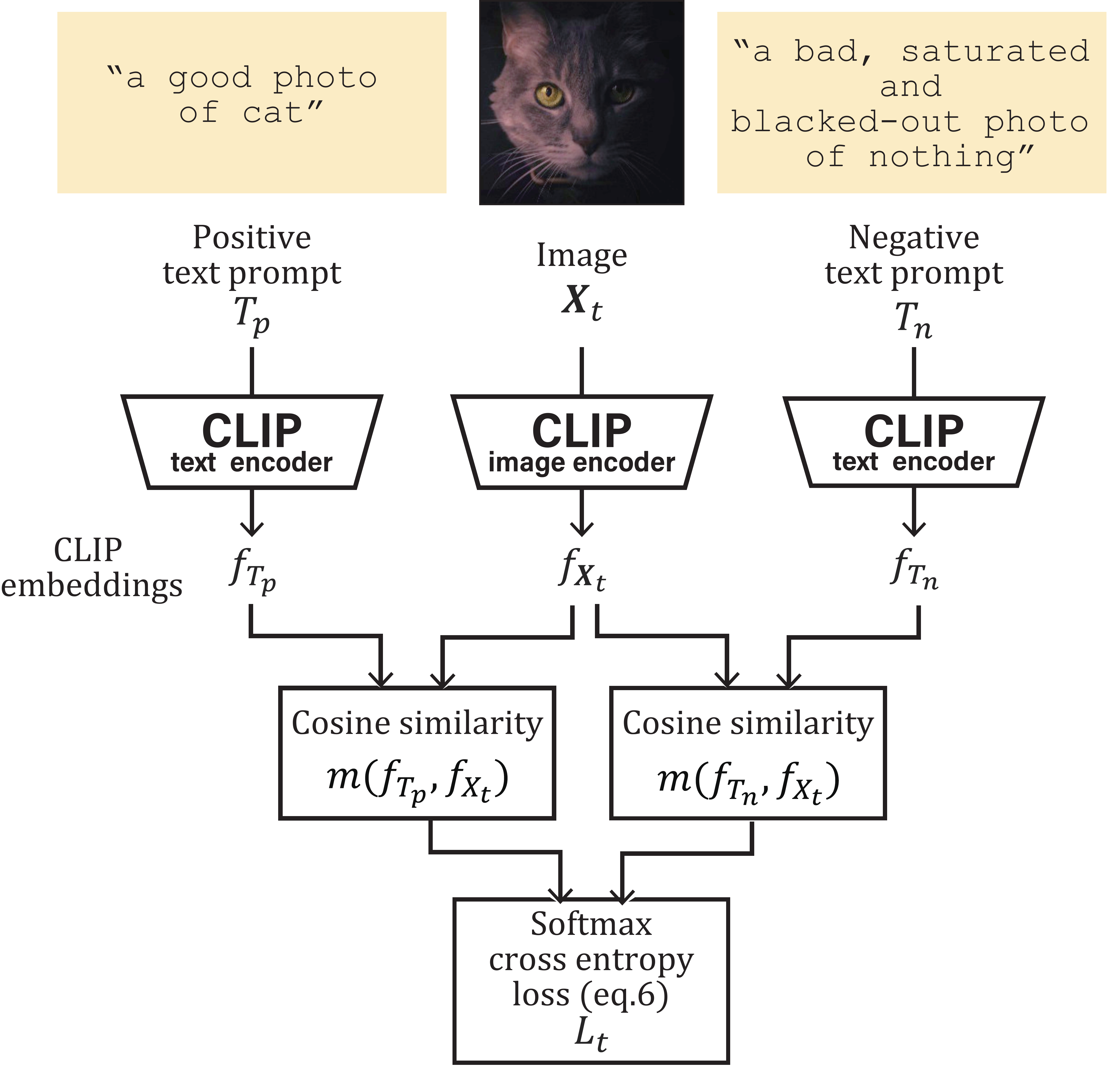}
    \\
     (a)&(b)
\end{tabular}
\caption{Training framework of our CURVE method. (a) The exploration flow of SAC. The obtained state, action and reward $[{\bf s}_t, {\bf a}_t, r_t, {\bf s}_{t+1}, {\bf a}_{t+1}]$ are stored in the replay buffer and used to update the policy network $\pi_\phi$ and Q network $Q_\theta$. The highlighted components (red and blue) are used for updating each network. (b) The CLIP-based reward calculation process.}
    \label{fig:train_loss}
\end{figure*}

\begin{figure*}
\centering
    \includegraphics[width=0.9\linewidth]{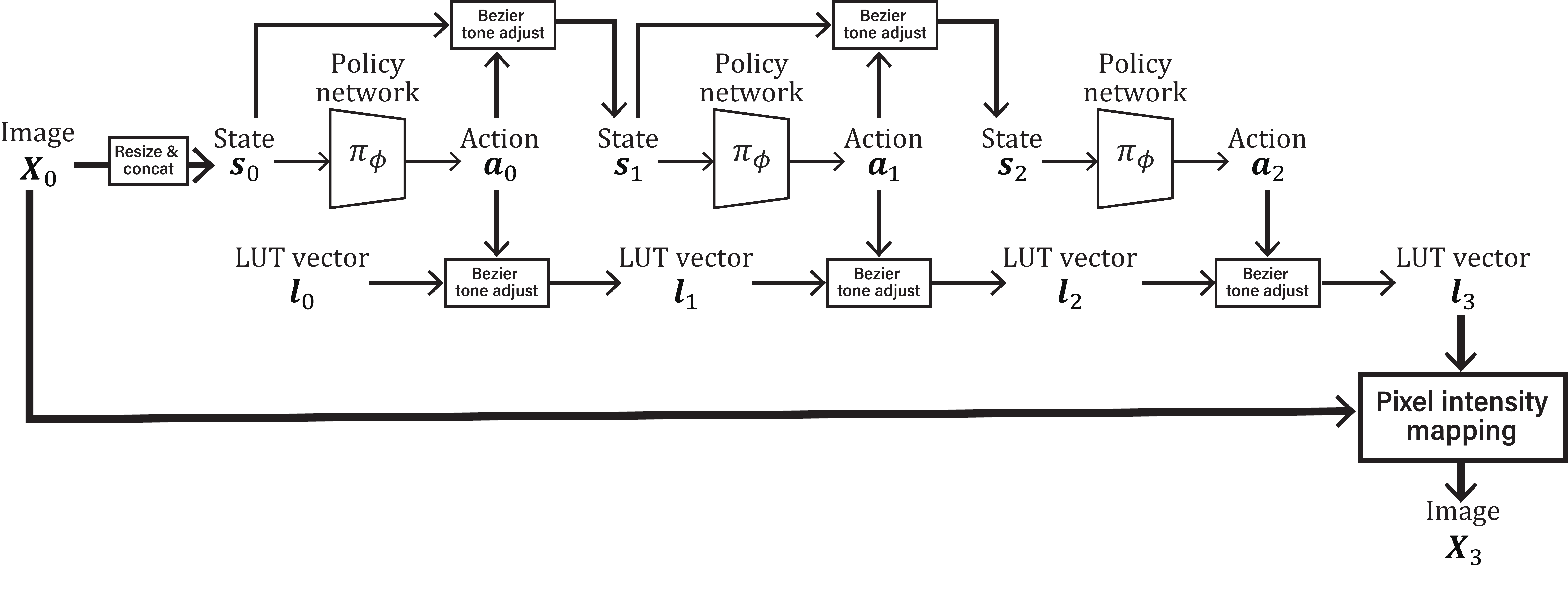}
\caption{Inference (testing) framework of our CURVE method}
    \label{fig:testing}
\end{figure*}

\newpage
\def\ims{0.31\linewidth}

\begin{figure*}
\centering
\begin{tabular}{ccc}
    \includegraphics[width=\ims]{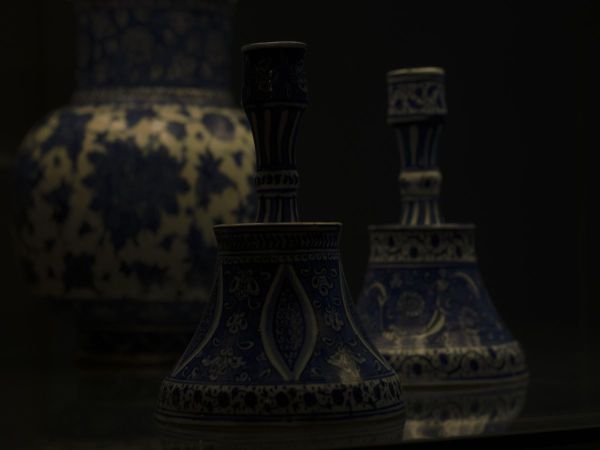}&
    \includegraphics[width=\ims]{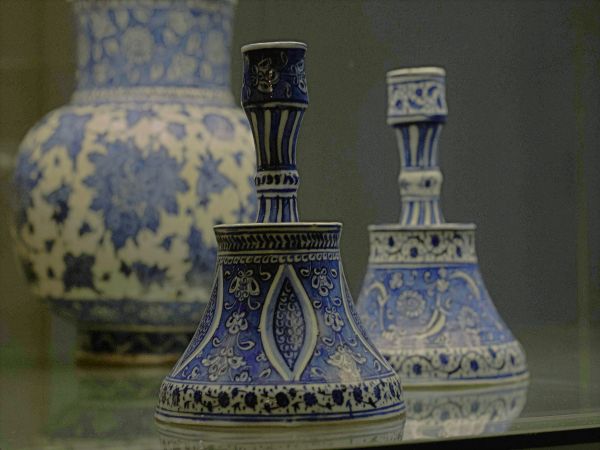}&
    \includegraphics[width=\ims]{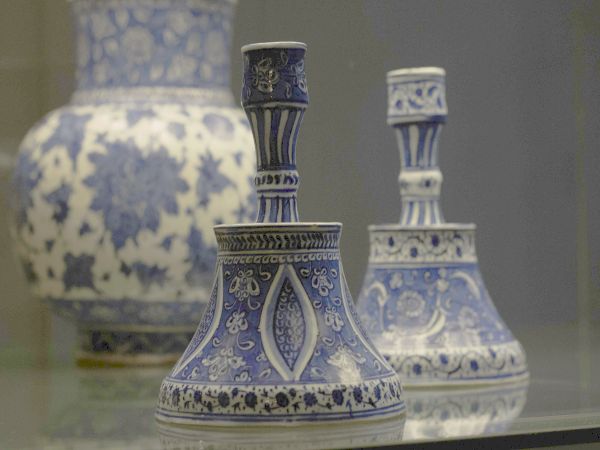}
     \\
    \includegraphics[width=\ims]{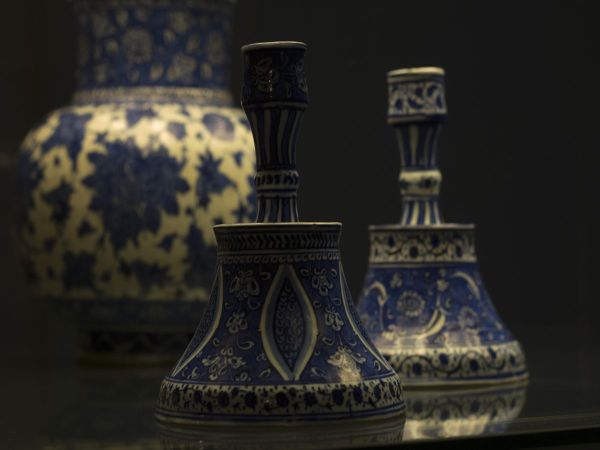}&
    \includegraphics[width=\ims]{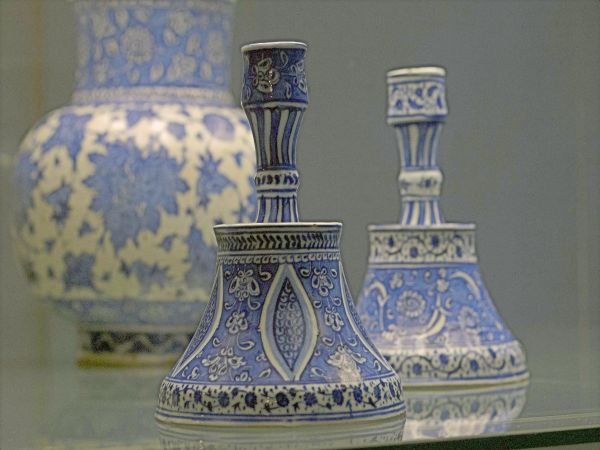}&
    \includegraphics[width=\ims]{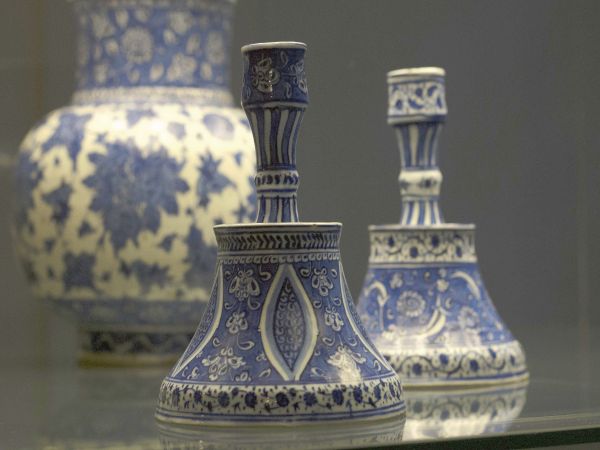}
     \\
     \includegraphics[width=\ims]{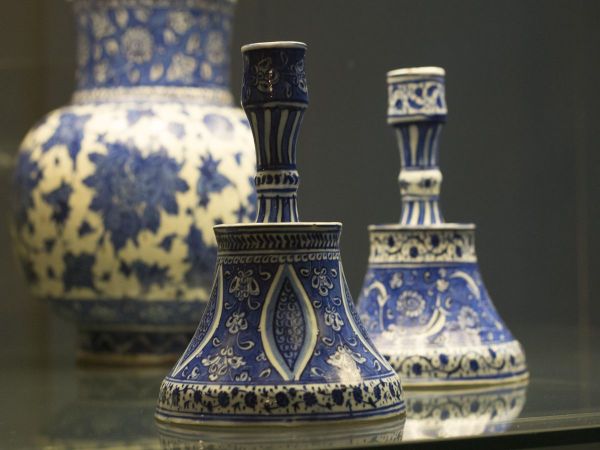}&
    \includegraphics[width=\ims]{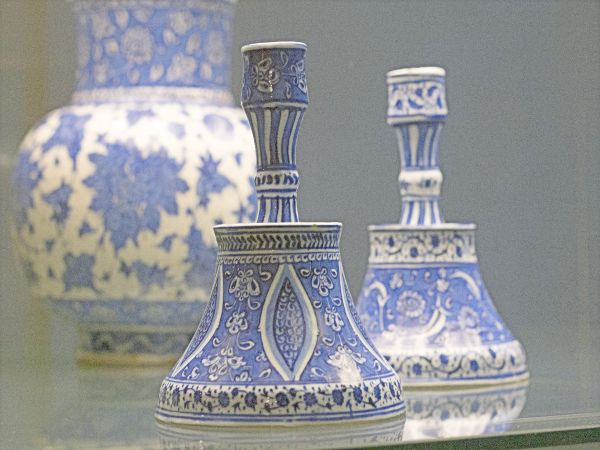}&
    \includegraphics[width=\ims]{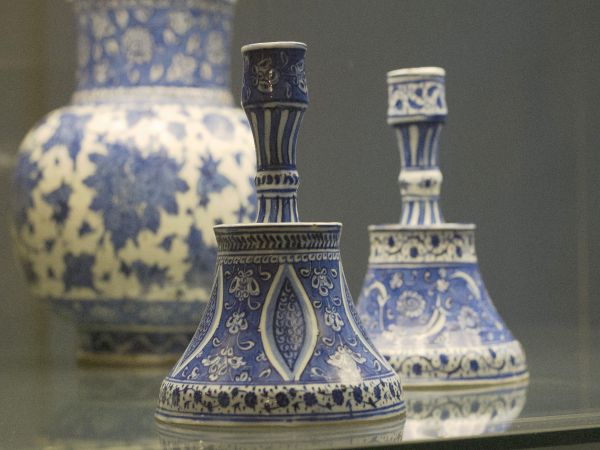}
     \\
     \includegraphics[width=\ims]{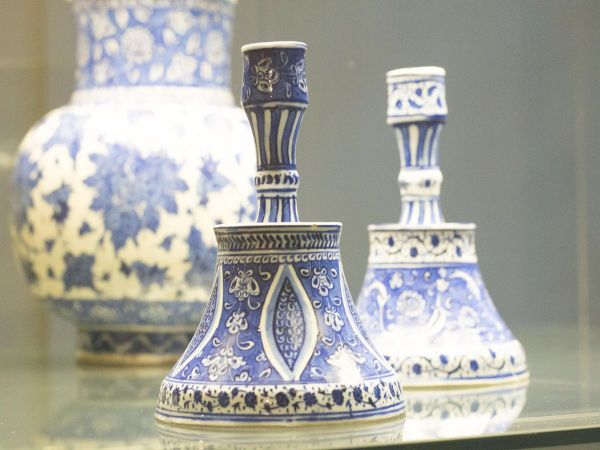}&
    \includegraphics[width=\ims]{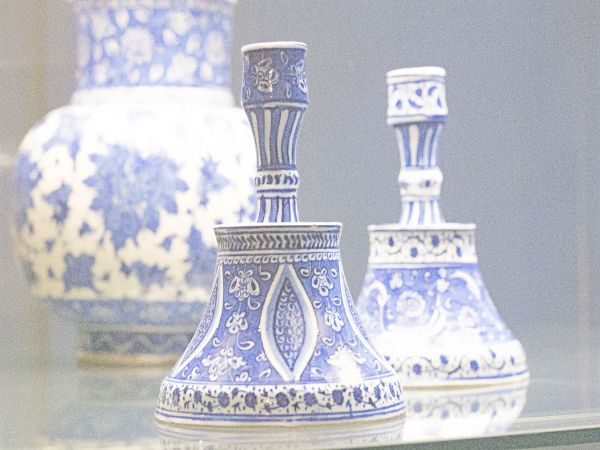}&
    \includegraphics[width=\ims]{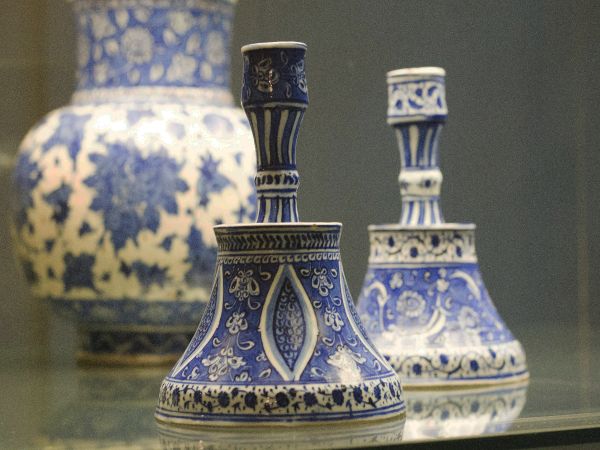}
     \\
     Original(input)&Zero-DCE&CURVE(ours)
\end{tabular}
\caption{Enhancement results on multi-exposure images from the SICE Part 2 dataset. Left: Input multi-exposure images. Middle: Results of Zero-DCE. Right: Results of our proposed CURVE.}
    \label{fig:sice-1}
\end{figure*}
\def\ims{0.31\linewidth}

\begin{figure*}
\centering
\begin{tabular}{ccc}
    \includegraphics[width=\ims]{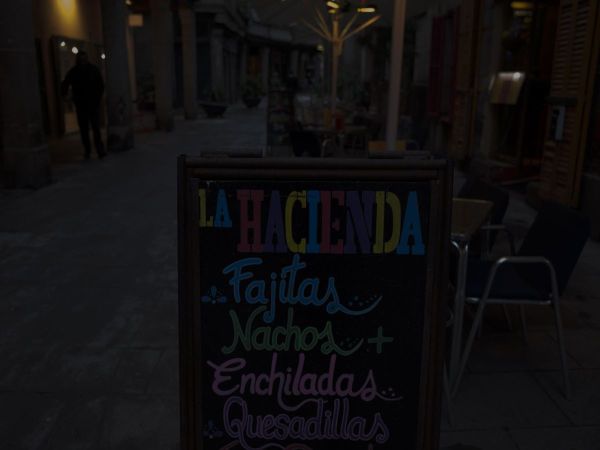}&
    \includegraphics[width=\ims]{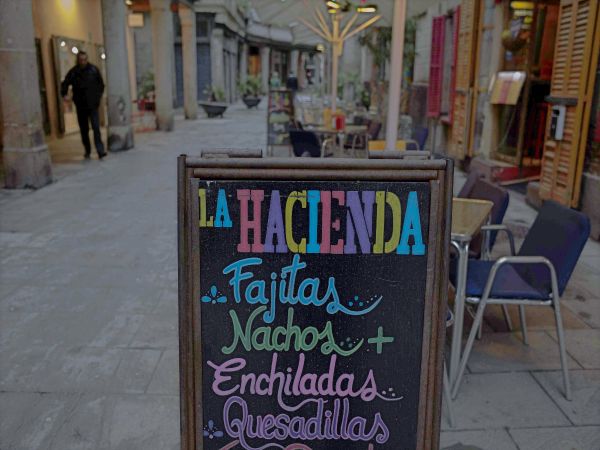}&
    \includegraphics[width=\ims]{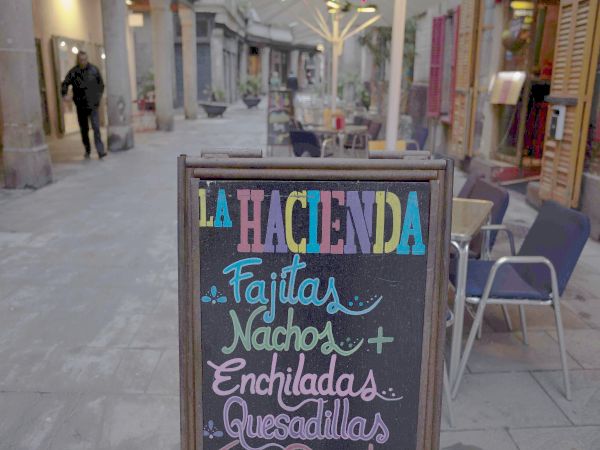}
     \\
    \includegraphics[width=\ims]{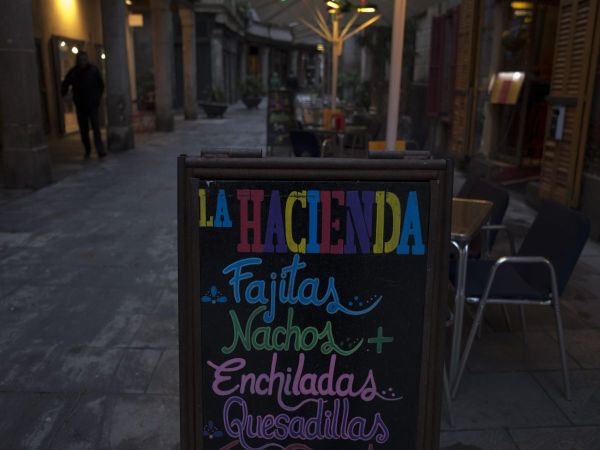}&
    \includegraphics[width=\ims]{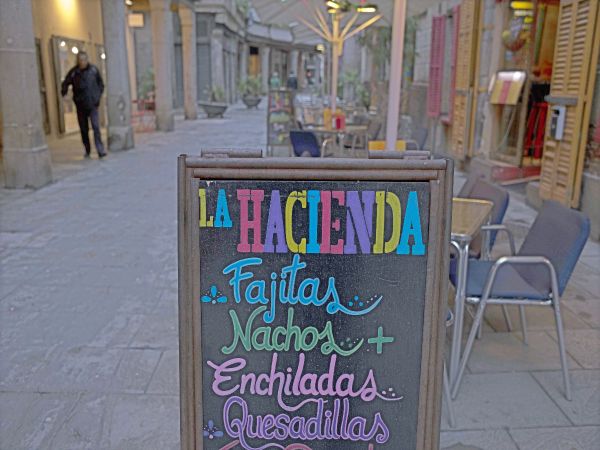}&
    \includegraphics[width=\ims]{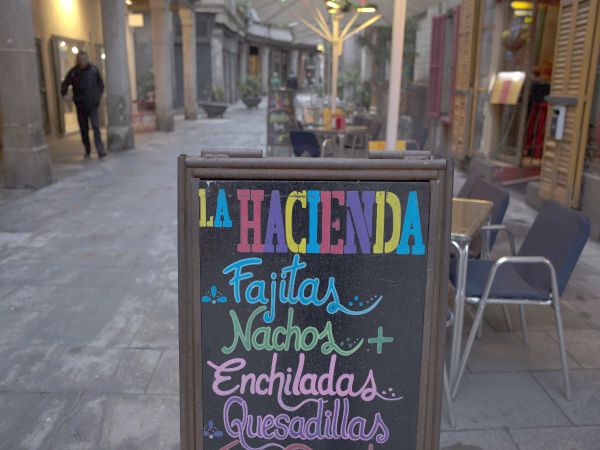}
     \\
     \includegraphics[width=\ims]{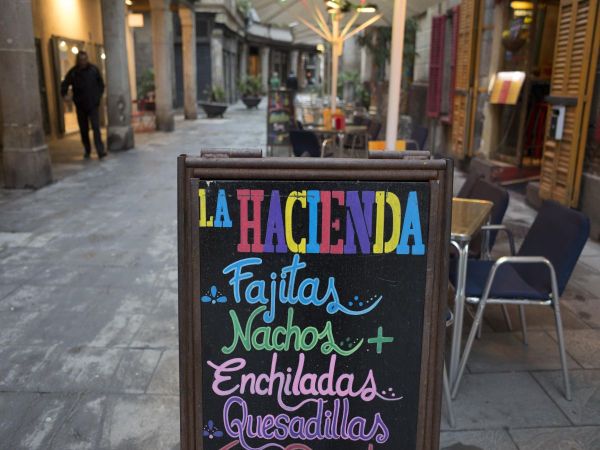}&
    \includegraphics[width=\ims]{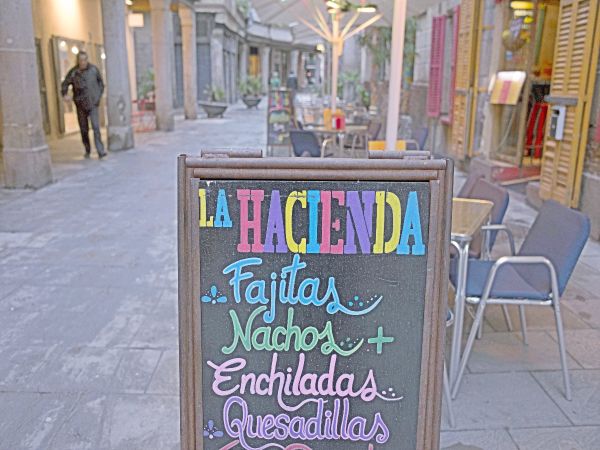}&
    \includegraphics[width=\ims]{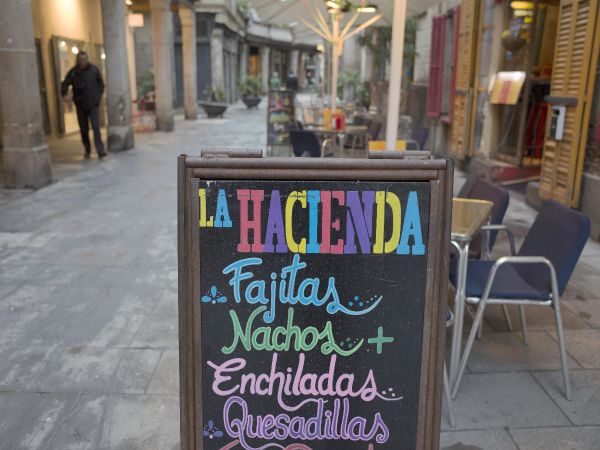}
     \\
     \includegraphics[width=\ims]{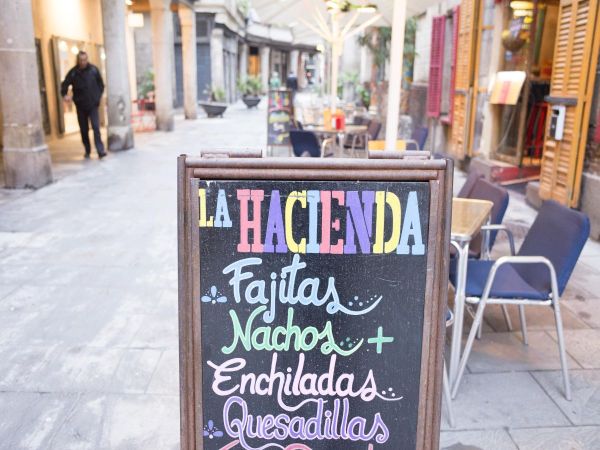}&
    \includegraphics[width=\ims]{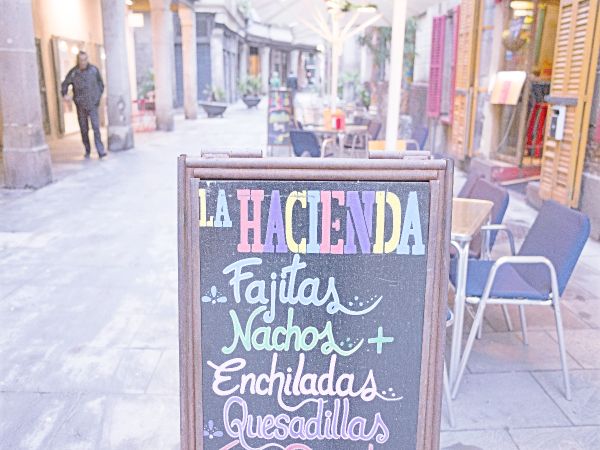}&
    \includegraphics[width=\ims]{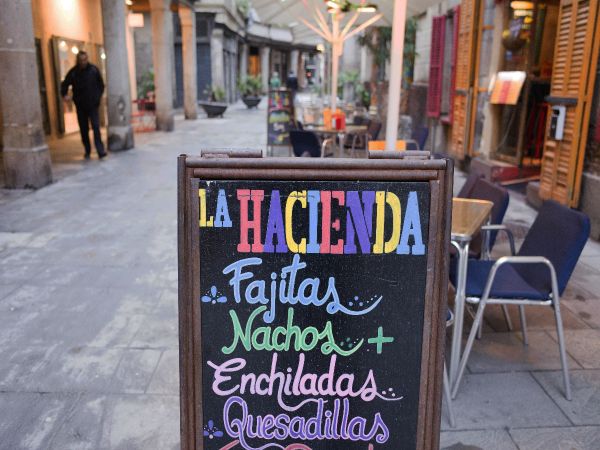}
     \\
     Original(input)&Zero-DCE&CURVE(ours)
\end{tabular}
\caption{Enhancement results on multi-exposure images from the SICE Part 2 dataset. Left: Input multi-exposure images. Middle: Results of Zero-DCE. Right: Results of our proposed CURVE.}
    \label{fig:sice-2}
\end{figure*}
\def\ims{0.31\linewidth}

\begin{figure*}
\centering
\begin{tabular}{ccc}
    \includegraphics[width=\ims]{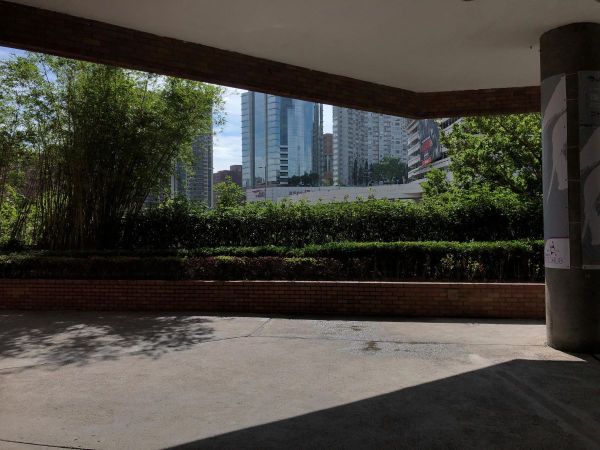}&
    \includegraphics[width=\ims]{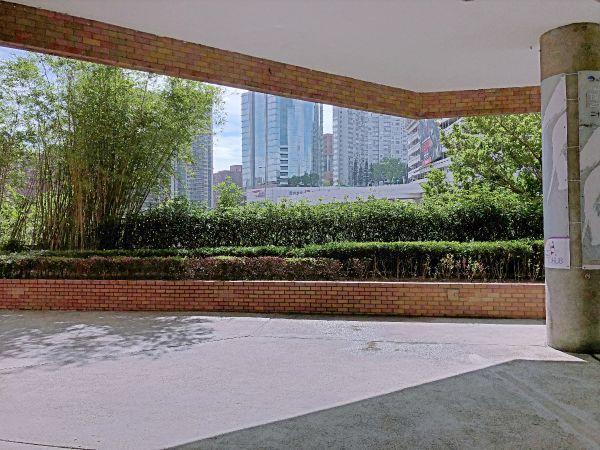}&
    \includegraphics[width=\ims]{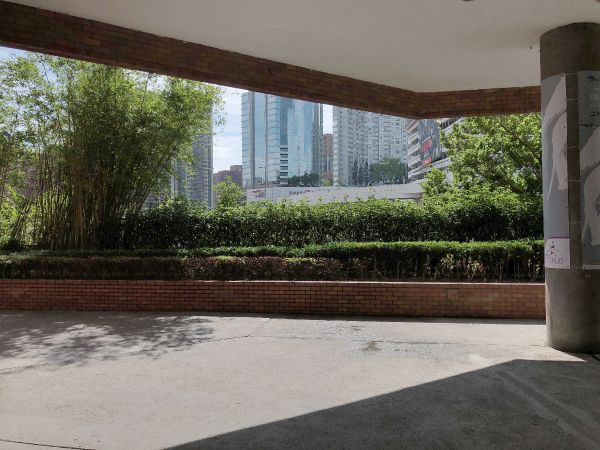}
     \\
    \includegraphics[width=\ims]{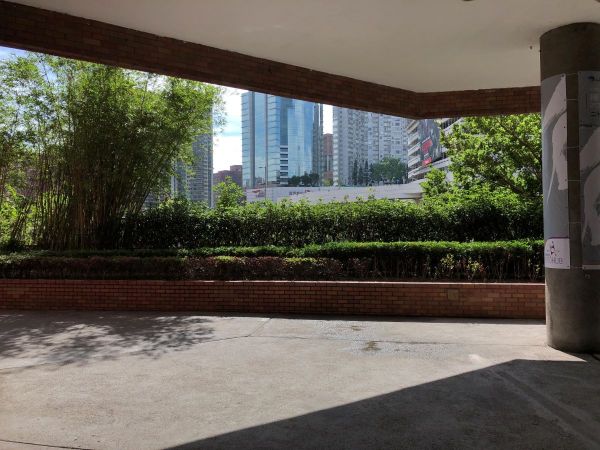}&
    \includegraphics[width=\ims]{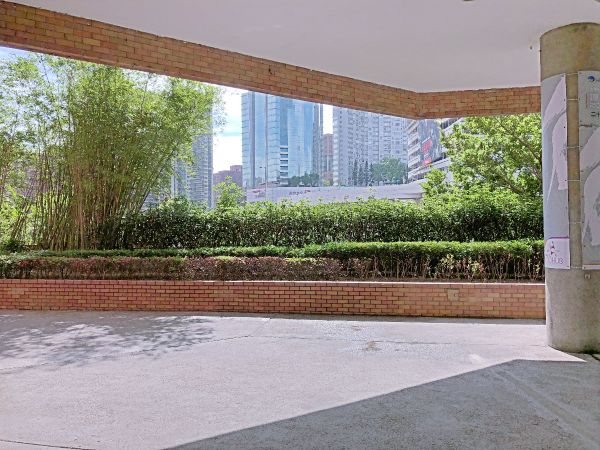}&
    \includegraphics[width=\ims]{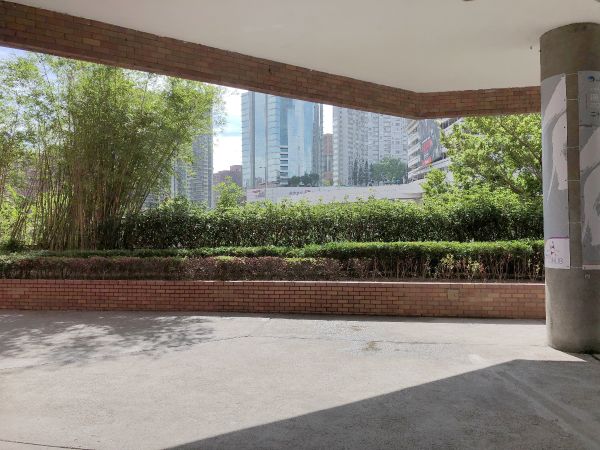}
     \\
     \includegraphics[width=\ims]{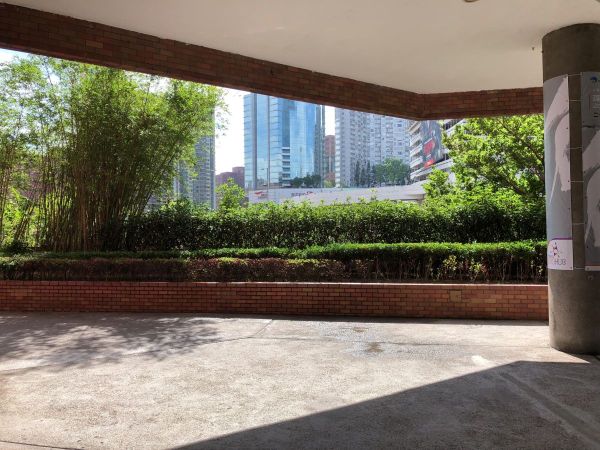}&
    \includegraphics[width=\ims]{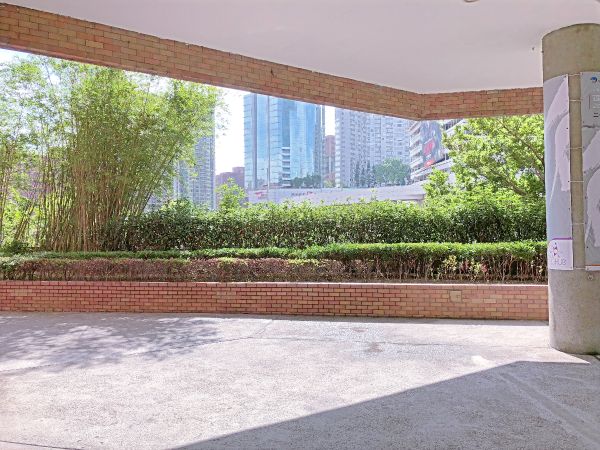}&
    \includegraphics[width=\ims]{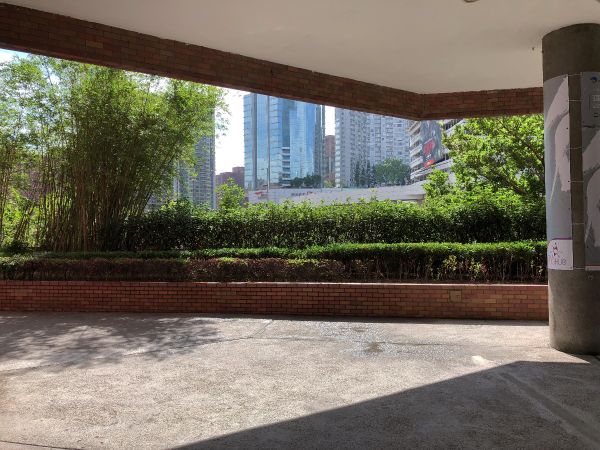}
     \\
     \includegraphics[width=\ims]{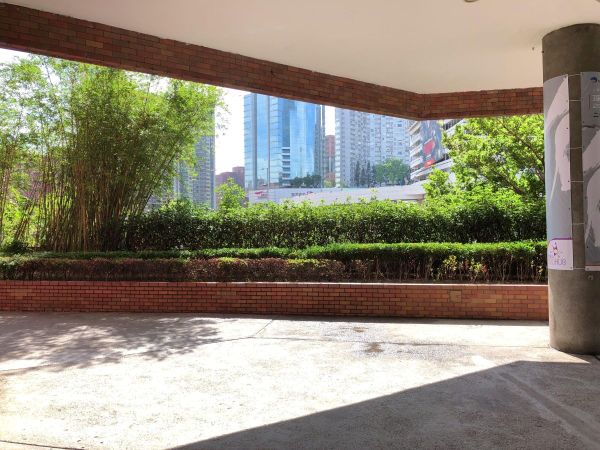}&
    \includegraphics[width=\ims]{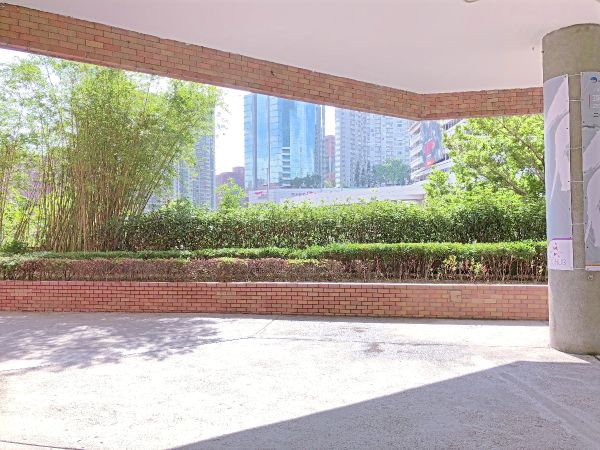}&
    \includegraphics[width=\ims]{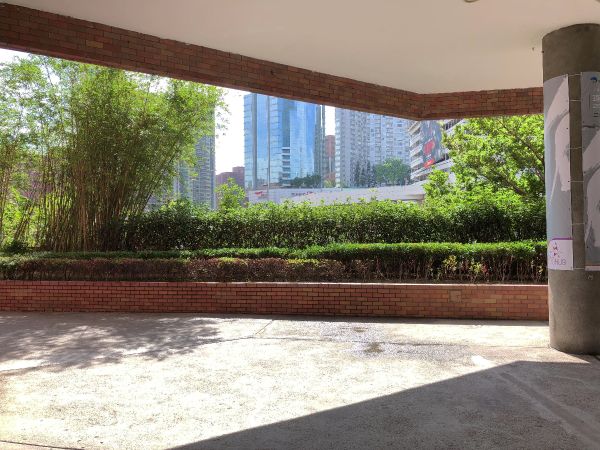}
     \\
      \includegraphics[width=\ims]{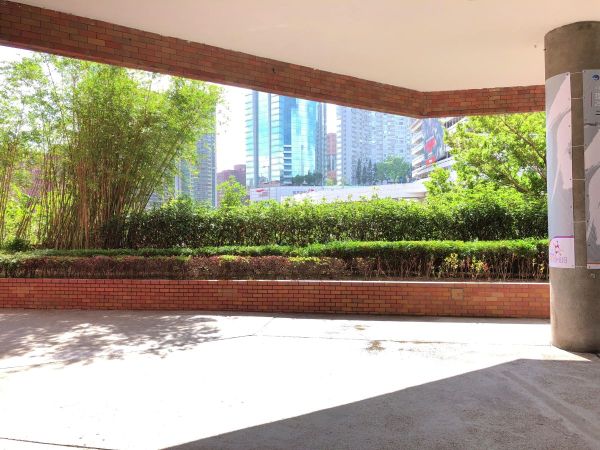}&
    \includegraphics[width=\ims]{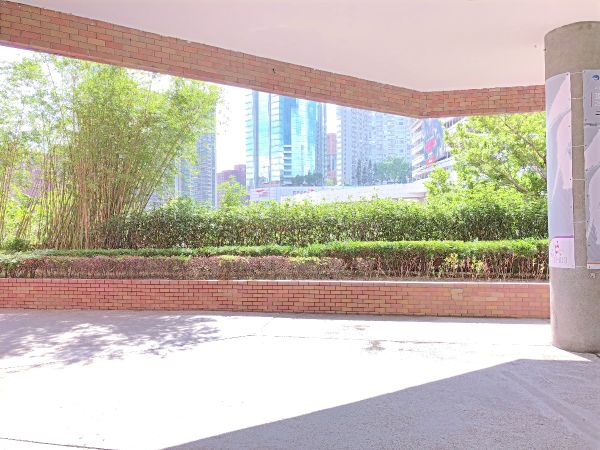}&
    \includegraphics[width=\ims]{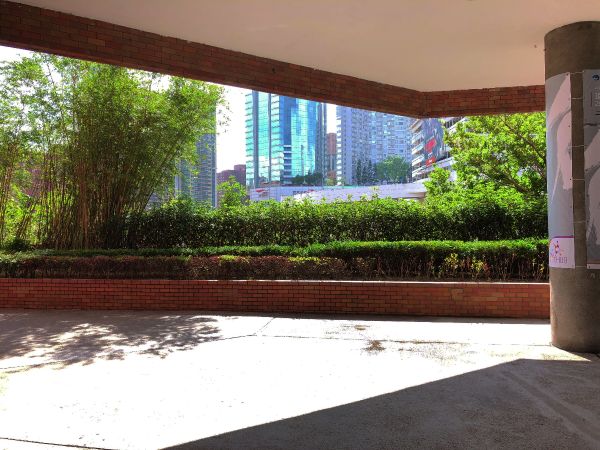}
     \\
     Original(input)&Zero-DCE&CURVE(ours)
\end{tabular}
\caption{Enhancement results on multi-exposure images from the SICE Part 2 dataset. Left: Input multi-exposure images. Middle: Results of Zero-DCE. Right: Results of our proposed CURVE.}
    \label{fig:sice-3}
\end{figure*}

\def\ims{0.31\linewidth}

\begin{figure*}
\centering
\begin{tabular}{cc}
     \includegraphics[width=\ims]{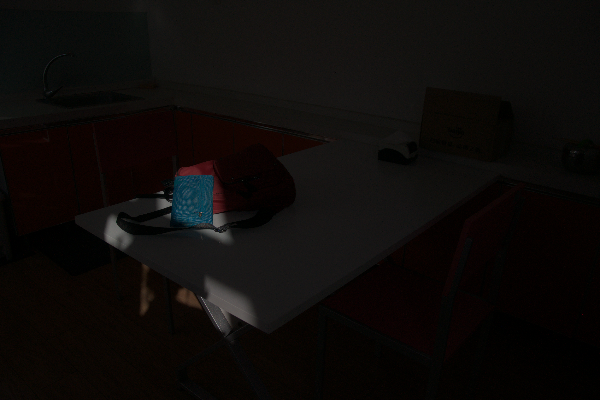}&
     \includegraphics[width=\ims]{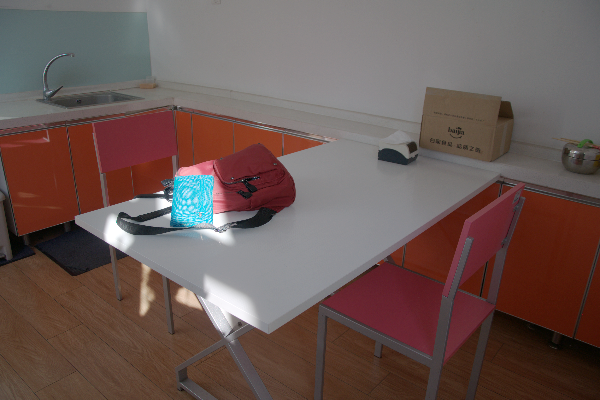}\\
     Original(input)& GT\\
     \includegraphics[width=\ims]{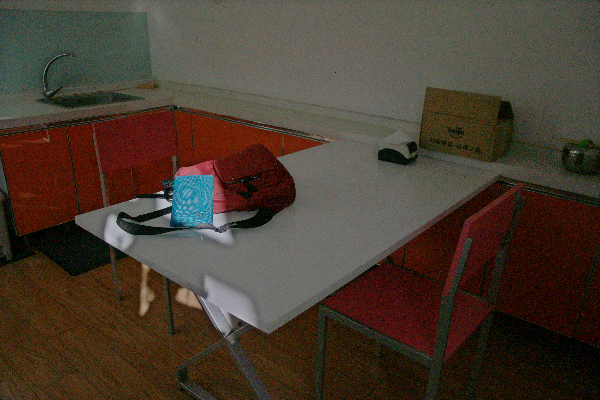}&
     \includegraphics[width=\ims]{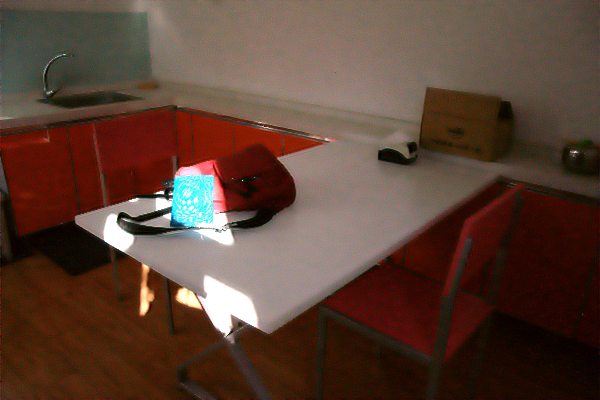}\\
     Zero-DCE&RUAS\\
     \includegraphics[width=\ims]{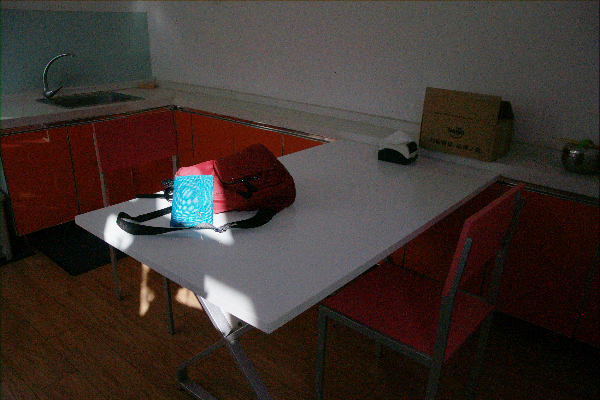}&
     \includegraphics[width=\ims]{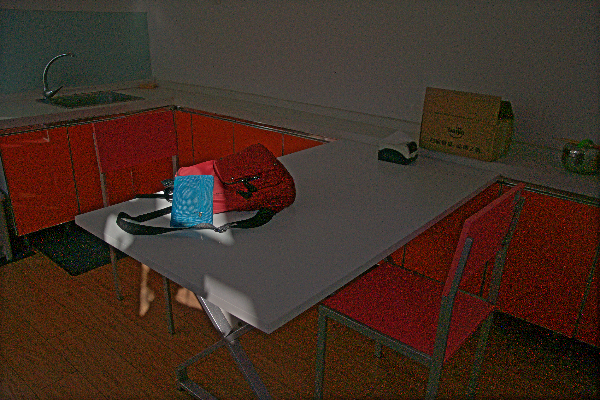}\\
     SCI&CLIP-LIT\\
     \includegraphics[width=\ims]{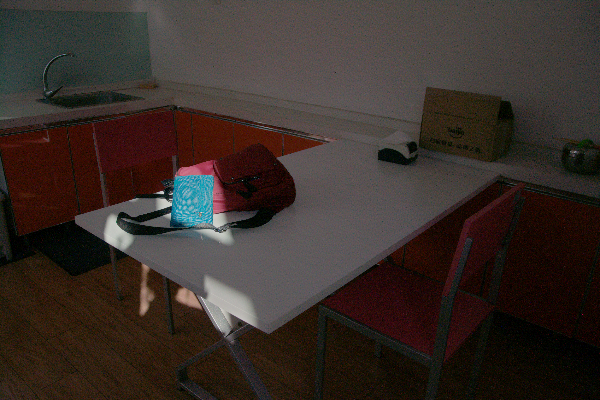}&
     \includegraphics[width=\ims]{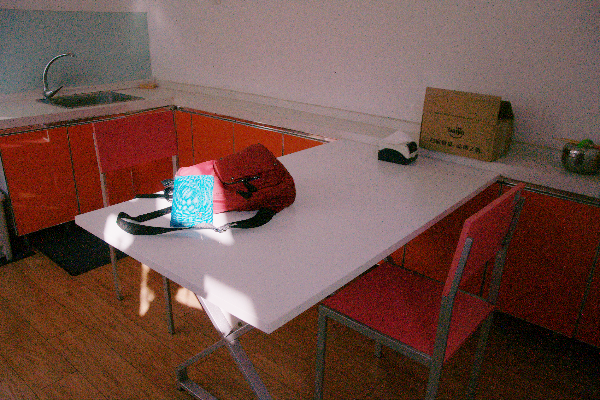}\\
     Morawski et al.&ReLLIE\\
     \includegraphics[width=\ims]{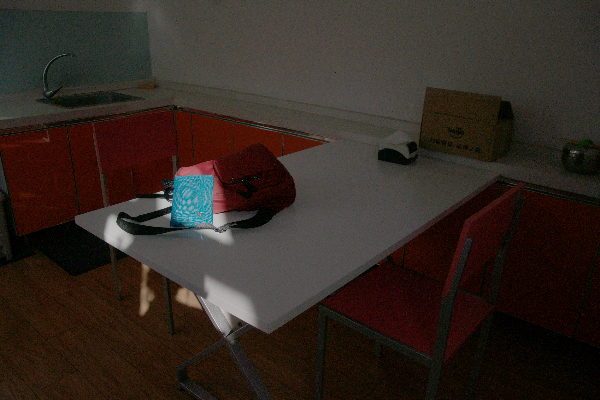}&
     \includegraphics[width=\ims]{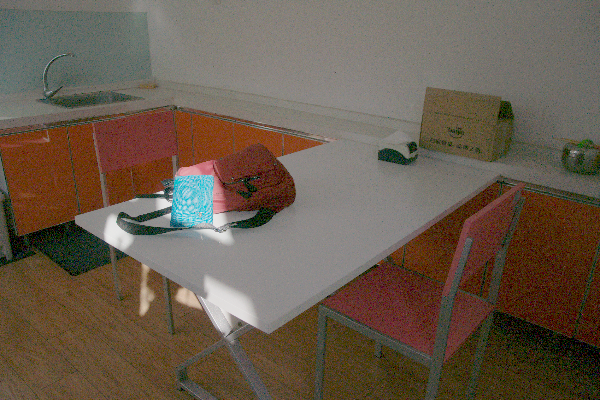}\\
     train-by-loss&CURVE(ours)
\end{tabular}
\caption{Enhancement results of our experiments on low-light images from the LoLv2Real dataset. The top row shows the input low-light image and ground truth (GT). Rows 2-5 show the results of six conventional zero-reference LLIE methods, an ablation study (train-by-loss), and our proposed CURVE.}
    \label{fig:lolv2-1}
\end{figure*}
\def\ims{0.31\linewidth}

\begin{figure*}
\centering
\begin{tabular}{cc}
     \includegraphics[width=\ims]{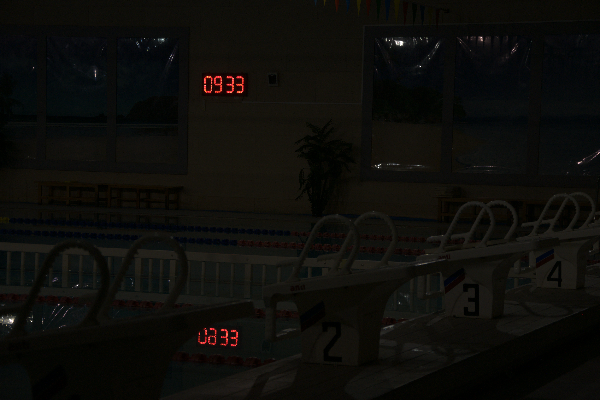}&
     \includegraphics[width=\ims]{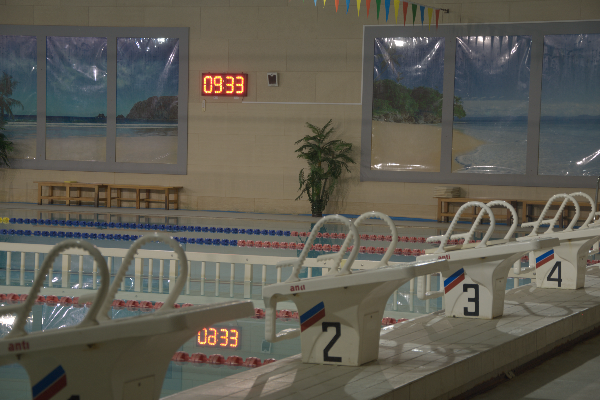}\\
     Original(input)& GT\\
     \includegraphics[width=\ims]{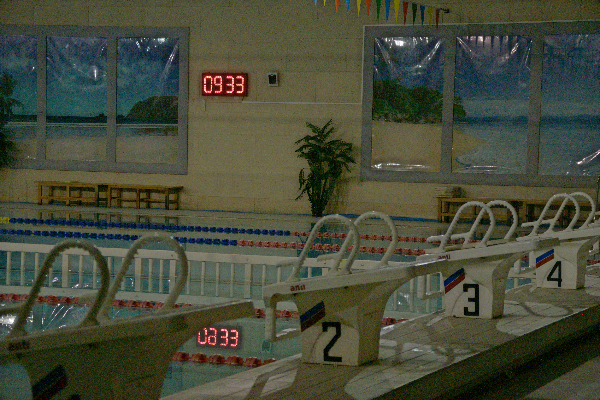}&
     \includegraphics[width=\ims]{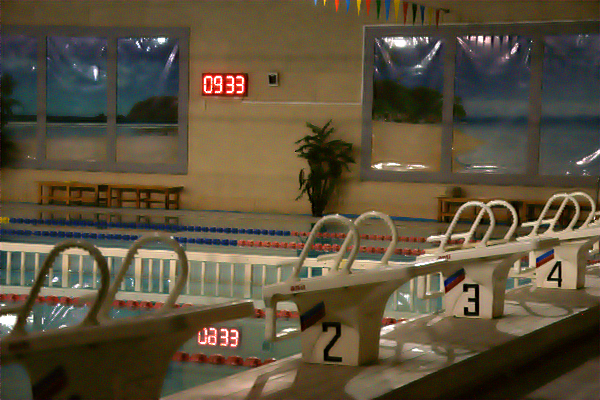}\\
     Zero-DCE&RUAS\\
     \includegraphics[width=\ims]{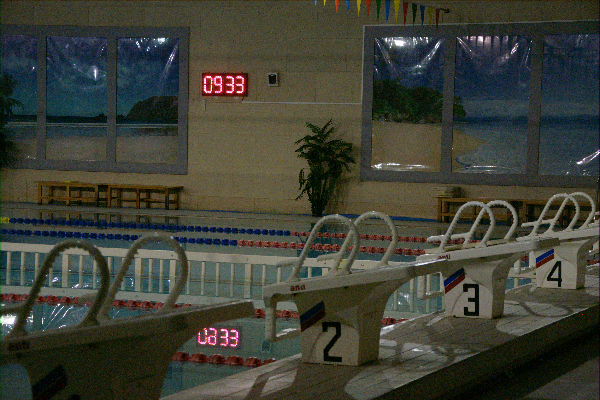}&
     \includegraphics[width=\ims]{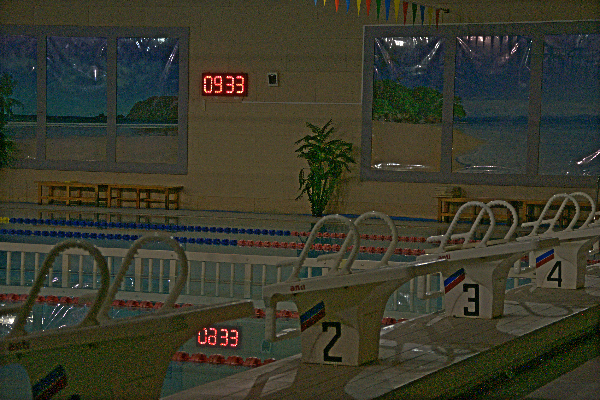}\\
     SCI&CLIP-LIT\\
     \includegraphics[width=\ims]{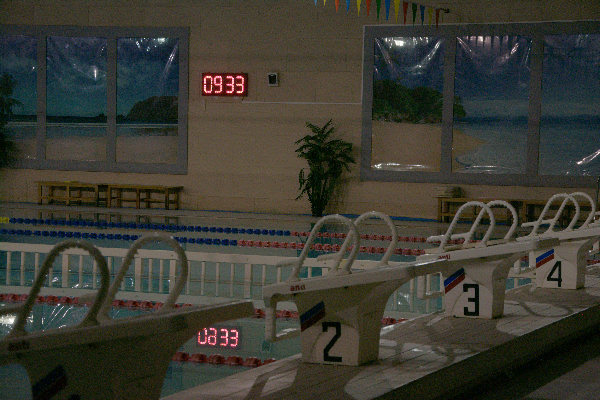}&
     \includegraphics[width=\ims]{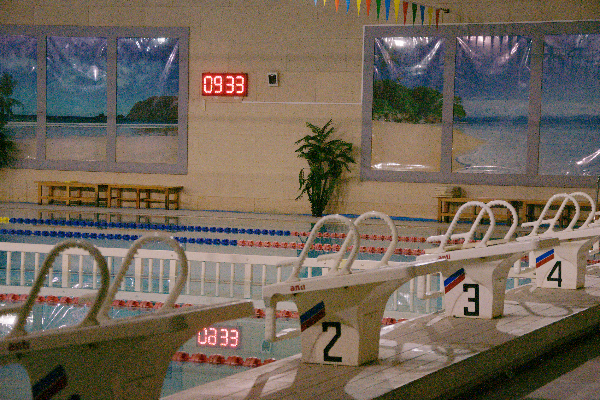}\\
     Morawski et al.&ReLLIE\\
     \includegraphics[width=\ims]{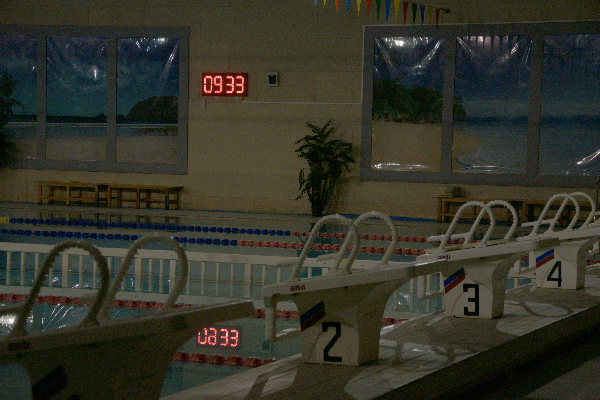}&
     \includegraphics[width=\ims]{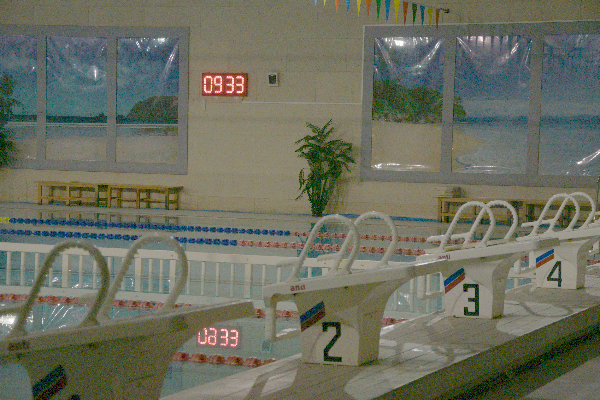}\\
     train-by-loss&CURVE(ours)
\end{tabular}
\caption{Enhancement results of our experiments on low-light images from the LoLv2Real dataset. The top row shows the input low-light image and ground truth (GT). Rows 2-5 show the results of six conventional zero-reference LLIE methods, an ablation study (train-by-loss), and our proposed CURVE.}
    \label{fig:lolv2-2}
\end{figure*}

\end{document}